\documentclass{article}

    \usepackage[nonatbib,preprint]{neurips_2021}

\usepackage[utf8]{inputenc} 
\usepackage[T1]{fontenc}    
\usepackage{hyperref}       
\usepackage{url}            
\usepackage{booktabs}       
\usepackage{amsfonts}       
\usepackage{nicefrac}       
\usepackage{microtype}      
\usepackage{amsmath,amssymb}
\usepackage{graphicx}
\usepackage{xcolor}
\usepackage{tabularx}
\usepackage{float}
\usepackage{wrapfig}
\usepackage{multicol}
\usepackage{multirow}
\usepackage{enumitem}
\usepackage{cite}  

\newcommand\etc{etc\@ifnextchar.{}{.\@}}
\newcommand\eg{e.g., \@}
\newcommand\ie{i.e., \@}
\newcommand\vs{vs. \@}
\newcommand\wrt{w.r.t. \@}
\definecolor{BPTT}{RGB}{134, 16, 1}
\definecolor{CBPTT}{RGB}{241, 133, 0}
\definecolor{CRBP}{RGB}{22, 79, 134}
\definecolor{RBP}{RGB}{183, 76, 187}

\newcommand{\rb}{\right]}
\newcommand{\lb}{\left[}

\title{Tracking Without Re-recognition in Humans and Machines}

\author{%
  Drew Linsley$^{*1}$, \hfill Girik Malik$^{*2}$, \hfill Junkyung Kim$^3$, \vspace{-4mm} \AND
  Lakshmi N Govindarajan$^1$, \hfill Ennio Mingolla$^{\dagger2}$, \hfill Thomas Serre$^{\dagger1}$ \\
  \texttt{\{drew\_linsley,lakshmi\_govindarajan,thomas\_serre\}@brown.edu} \\
  \texttt{\{malik.gi,e.mingolla\}@northeastern.edu} \\
  \texttt{junkyung@deepmind.com} \\
}

\begin{document}

\maketitle

\begin{abstract}
Imagine trying to track one particular fruitfly in a swarm of hundreds. Higher biological visual systems have evolved to track moving objects by relying on both appearance and motion features. We investigate if state-of-the-art deep neural networks for visual tracking are capable of the same. For this, we introduce \textit{PathTracker}, a synthetic visual challenge that asks human observers and machines to track a target object in the midst of identical-looking ``distractor'' objects. While humans effortlessly learn \textit{PathTracker} and generalize to systematic variations in task design, state-of-the-art deep networks struggle. To address this limitation, we identify and model circuit mechanisms in biological brains that are implicated in tracking objects based on motion cues. When instantiated as a recurrent network, our circuit model learns to solve \textit{PathTracker} with a robust visual strategy that rivals human performance and explains a significant proportion of their decision-making on the challenge. We also show that the success of this circuit model extends to object tracking in natural videos. Adding it to a transformer-based architecture for object tracking builds tolerance to visual nuisances that affect object appearance, resulting in a new state-of-the-art performance on the large-scale TrackingNet object tracking challenge. Our work highlights the importance of building artificial vision models that can help us better understand human vision and improve computer vision.
\end{abstract}

\section{Introduction}
\begin{wrapfigure}[20]{r}{0.6\textwidth}\vspace{-0mm}
  \centering
    \includegraphics[width=0.6\textwidth]{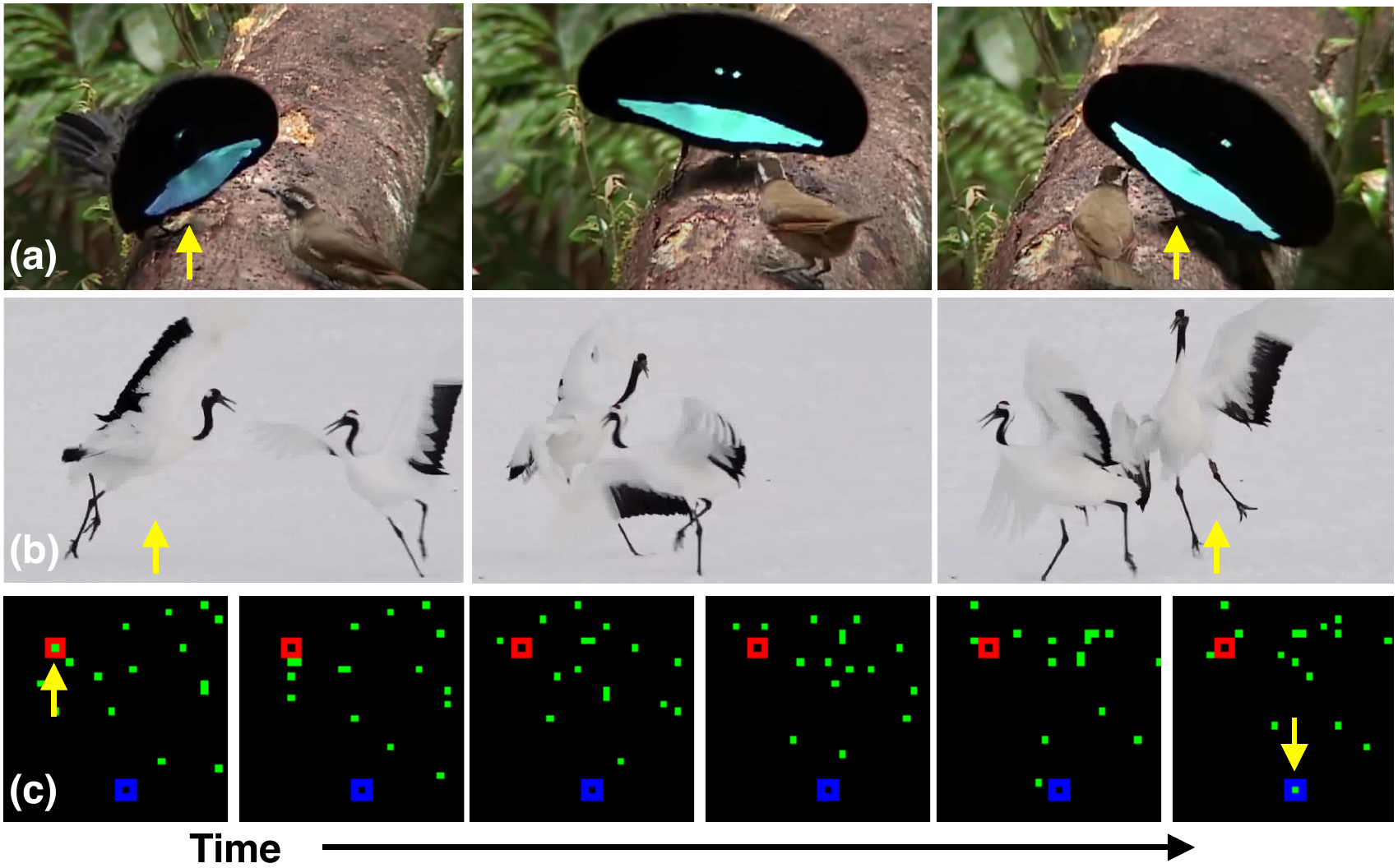}\vspace{-4mm}
  \caption{The appearance of objects makes them (\textit{a}) easy or (\textit{b}) hard to track. We introduce the \textit{PathTracker} Challenge (\textit{c}), which asks observers to track a particular green dot as it travels from the red-to-blue markers, testing object tracking when re-recognition is impossible.}\label{fig:teaser}
\end{wrapfigure}

\footnotetext[1]{$^\dagger$These authors contributed equally to this work.}
\footnotetext{$^{1}$Carney Institute for Brain Science, Brown University, Providence, RI}
\footnotetext{$^{2}$Northeastern University, Boston, MA}
\footnotetext{$^{3}$DeepMind, London, UK}
Lettvin and colleagues~\cite{Lettvin1959-ha} presciently noted, ``The frog does not seem to see or, at any rate, is not concerned with the detail of stationary parts of the world around him. He will starve to death surrounded by food if it is not moving.'' Object tracking is fundamental to survival, and higher biological visual systems have evolved the capacity for two distinct and complementary strategies to do it. Consider Figure~\ref{fig:teaser}: can you track the object labeled by the yellow arrow from left-to-right? The task is trivial when there are ``bottom-up'' cues for object appearance, like color, which make it possible to ``re-recognize'' the target in each frame (Fig.~\ref{fig:teaser}a). On the other hand, the task is more challenging when objects cannot be discriminated by their appearance. In this case integration of object motion over time is necessary for tracking (Fig.~\ref{fig:teaser}b). Humans are capable of tracking objects by their motion when appearance is uninformative~\cite{Pylyshyn1988-pi,Blaser2000-xz}, but it is unclear if the current generation of neural networks for video analysis and tracking can do the same. To address this question we introduce \textit{PathTracker}, a synthetic challenge for object tracking without re-recognition (Fig.~\ref{fig:teaser}c).

Leading models for video analysis rely on object classification pre-training. This gives them access to rich semantic representations that have supported state-of-the-art performance on a host of tasks, from action recognition to object tracking~\cite{carreira2017,Bertasius2021-dk,Wang2021-no}. As object classification models have improved, so too have the video analysis models that depend on them. This trend in model development has made it unclear if video analysis models are effective at learning tasks when appearance is uninformative. The importance of diverse visual strategies has been highlighted by synthetic challenges like \textit{Pathfinder},a visual reasoning task that asks observers to trace long paths embedded in a static cluttered display~\cite{Linsley2018-ls,Kim2020-yw}. \textit{Pathfinder} tests object segmentation when appearance cues like category or shape are missing. While humans can easily solve it ~\cite{Kim2020-yw}, feedforward neural networks struggle, including state-of-the-art vision transformers~\cite{Tay2020-ni,Kim2020-yw,Linsley2018-ls}. Importantly, models that learn an appropriate visual strategy for \textit{Pathfinder} are also quicker learners and better at generalization for object segmentation in natural images~\cite{Linsley2020-en,Linsley2020-ua}. Our \textit{PathTracker} challenge extends this line of work into video by posing an object tracking problem where the target can be tracked by motion and spatiotemporal continuity, not category or appearance.

\paragraph{Contributions.} 
Humans effortlessly solve our novel \textit{PathTracker} challenge. A variety of state-of-the-art models for object tracking and video analysis do not.
\begin{itemize}[leftmargin=*]\vspace{-2mm}
    \item We find that neural architectures including R3D~\cite{Tran2017-fg} and state-of-the-art transformer-based TimeSformers~\cite{Bertasius2021-dk} are strained by long \textit{PathTracker} videos. Humans, on the other hand, are far more effective at solving these long  \textit{PathTracker} videos. 
    \item We describe a solution to \textit{PathTracker}: a recurrent model inspired by primate neural circuitry involved in object tracking, whose decisions that are strongly correlated with those of humans.  
    \item These same circuit mechanisms improve object tracking in natural videos through a motion-based strategy that builds tolerance to changes in target object appearance, resulting in the certified top score on TrackingNet~\cite{Muller2018-qn} at the time of this submission.
    \item We release all \textit{PathTracker} data, code, and human psychophysics at \url{http://bit.ly/InTcircuit} to spur interest in the challenge of tracking without re-recognition. 
\end{itemize}




\section{Related Work}

\paragraph{Shortcut learning and synthetic datasets} A byproduct of the great power of deep neural network architectures is their vulnerability to learning spurious correlations between inputs and labels. Perhaps because of this tendency, object classification models have trouble generalizing to novel contexts~\cite{Barbu2019-zq, Geirhos2020-nl}, and render idiosyncratic decisions that are inconsistent with humans~\cite{Ullman2016-ea,Linsley2017-qe,Linsley2019-bw}. Synthetic datasets are effective at probing this vulnerability because they make it possible to control spurious image/label correlations and fairly test the computational abilities of models. For example, the \textit{Pathfinder} challenge was designed to test if neural architectures can trace long curves despite gaps -- a visual computation associated with the earliest stages of visual processing in primates. That challenge identified diverging visual strategies between humans and transformers that are otherwise state of the art in natural image object recognition~\cite{Tay2020-ni,Dosovitskiy2020-if}. Other challenges like Bongard-LOGO~\cite{Nie2020-lx}, cABC~\cite{Kim2020-yw}, and PSVRT~\cite{Kim2018-ib} have highlighted limitations of leading neural network architectures that would have been difficult to identify using natural image benchmarks like ImageNet~\cite{Deng2009-jk}. These limitations have inspired algorithmic solutions based on neural circuits discussed in SI \S\ref{si_sec:related_work}.


\paragraph{Models for video analysis} A major leap in the performance of models for video analysis came from using networks which are pre-trained for object recognition on large image datasets~\cite{Carreira2017-ic}. The recently introduced TimeSformer~\cite{Bertasius2021-dk} achieved state-of-the-art performance with weights initialized from an image categorization transformer (ViT; \cite{Dosovitskiy2020-if}) that was pre-trained on ImageNet-21K. The story is similar in object tracking~\cite{Fiaz2018-zi}, where successful models rely on ``backbone'' feature extraction networks trained on ImageNet or Microsoft COCO~\cite{Lin2014-zk} for object recognition or segmentation~\cite{Bertasius2020-eo,Wang2021-no}. 



\begin{figure}[t]
\begin{center}\small
\includegraphics[width=.99\textwidth]{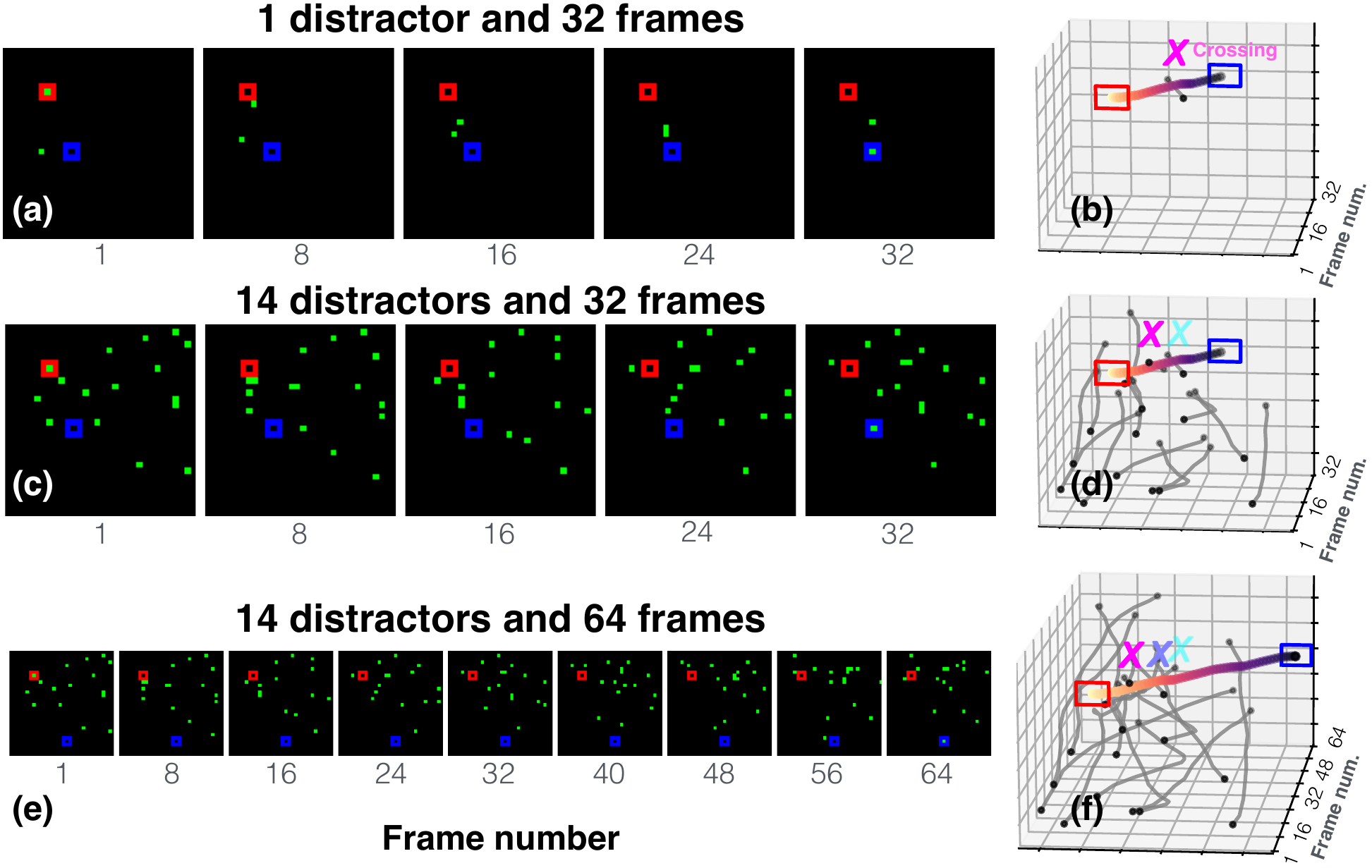}
\end{center}
\vspace{-4mm}
\caption{\textit{PathTracker} is a synthetic visual challenge that asks observers to watch a video clip and answer if a target dot starting in a red marker travels to a blue marker. The target dot is surrounded by identical ``distractor'' dots, each of which travels in a randomly generated and curved path. In positive examples, the target dot's path ends in the blue square. In negative examples, a ``distractor'' dot ends in the blue square. The challenge of the task is due to the identical appearance of target and distractor dots, which makes appearance-based tracking strategies ineffective. Moreover, the target dot can momentarily occupy the same location as a distractor when they cross each other's paths, making them impossible to individuate in that frame and compelling strategies like motion trajectory extrapolation or working memory to recover the target track. (\textit{b}) A 3D visualization of the video in (\textit{a}) depicts the trajectory of the target dot, traveling from red-to-blue markers. The target and distractor cross approximately half-way through the video. (\textit{c,d}) We develop versions of \textit{PathTracker} that test observer sensitivity to the number distractors and length of videos (\textit{e,f}). The number of distractors and video length interact to make it more likely for the target dot to cross a distractor in a video (compare the one X in \textit{b} \vs two in \textit{d} \vs three in \textit{f}; see SI \S\ref{si_sec:benchmark} for details).}\vspace{-4mm}
\label{fig:dataset}
\end{figure}\section{The \textit{PathTracker} Challenge}
\label{sec:challenge}

\paragraph{Overview} \textit{PathTracker} asks observers to decide whether or not a target dot reaches a goal location (Fig.~\ref{fig:dataset}). The target dot travels in the midst of a pre-specified number of distractors. All dots are identical, and the task is difficult because of this: (\textit{i}) observers cannot rely on appearance to track the target, and (\textit{ii}) the paths of the target and distractors can momentarily ``cross'' and occupy the same space, making them impossible to individuate in that frame and meaning that observers cannot only rely on target location to solve the task. This challenge is inspired by object tracking paradigms of cognitive psychology \cite{Pylyshyn1988-pi, Blaser2000-xz}, which suggest that humans might rely on mechanisms for motion perception, attention and working memory to solve a task like \textit{PathTracker}.

\begin{figure}[t]
\begin{center}\small
\includegraphics[width=.99\textwidth]{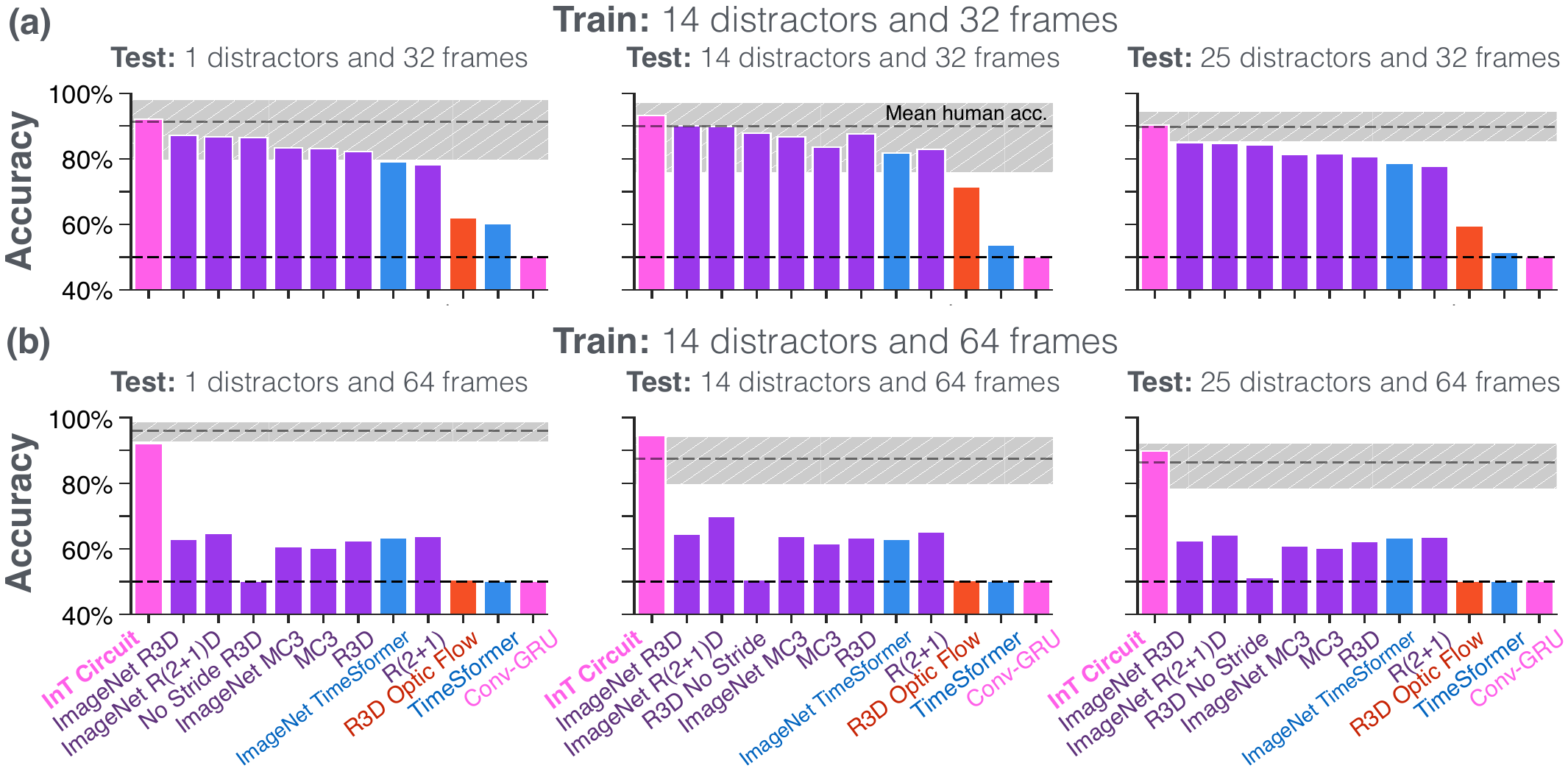}
\end{center}
\vspace{-4mm}
\caption{Model accuracy on the \textit{PathTracker} challenge. Video analysis models were trained to solve 32 (\textit{a}) and 64 frame (\textit{b}) versions of challenge, which featured the target object and 14 identical distractors. Models were tested on \textit{PathTracker} datasets with the same number of frames but 1, 14, or 25 distractors (left/middle/right). Grey hatched boxes denote 95\% bootstrapped confidence intervals for humans. Only our InT Circuit rivaled humans on each dataset.}\vspace{-4mm}
\label{fig:model_perf}
\end{figure}


The trajectories of target and distractor dots are randomly generated, and the target occasionally crosses distractors (Fig.~\ref{fig:dataset}). These object trajectories are smooth by design, giving the appearance of objects meandering through a scene, and the difference between the coordinates of any dot on successive frames is no more than 2 pixels with less than $20^\circ$ of angular displacement. In other words, dots never turn at acute angles. We develop different versions of \textit{Pathtracker} which we expect to be more or less difficult by adjusting the number of distractors and/or the length of videos. These variables change the expected number of times that distractors cross the target and the amount of time that observers must track the target (Fig.~\ref{fig:dataset}). To make the task as visually simple as possible and maximize contrast between dots and markers, the dots, start, and goal markers are placed on different channels in 32$\times$32 pixel three-channel images. Markers are stationary throughout each video and placed at random locations. Examples videos can be viewed at \url{http://bit.ly/InTcircuit}.

\paragraph{Human benchmark} We began by testing if humans can solve \textit{PathTracker}. We recruited 180 individuals using Amazon Mechanical Turk to participate in this study. Participants viewed \textit{PathTracker} videos and pressed a button on their keyboard to indicate if the target object or a distractor reached the goal. These videos were played in web browsers at 256$\times$256 pixels using HTML5, which helped ensure consistent framerates~\cite{Eberhardt2016-cw}. The experiment began with an 8 trial ``training'' stage, which familiarized participants with the goal of \textit{PathTracker}. Next, participants were tested on 72 videos. The experiment was not paced and lasted approximately 25 minutes, and participants were paid $\$8$ for their time. See \url{http://bit.ly/InTcircuit} and SI \S\ref{si_sec:benchmark} for an example and more details.

Participants were randomly entered into one of two experiments. In the first experiment, they were trained on the 32 frame and 14 distractor \textit{PathTracker}, and tested on 32 frame versions with 1, 14, or 25 distractors. In the second experiment, they were trained on the 64 frame and 14 distractor \textit{PathTracker}, and tested on 64 frame versions with 1, 14, or 25 distractors. All participants viewed unique videos to maximize our sampling over the different versions of \textit{PathTracker}. Participants were significantly above chance on all tested conditions of \textit{PathTracker} (\textit{p} $<$ 0.001, test details in SI \S\ref{si_sec:benchmark}). They also exhibited a significant negative trend in performance on the 64 frame datasets as the number of distractors increased ($t=-2.74$, $p < 0.01$). There was no such trend on the 32 frame datasets, and average accuracy between the two datasets was not significantly different. These results validate our initial design assumptions: humans can solve \textit{PathTracker}, and manipulating distractors and video length increases difficulty.

\section{Solving the \textit{PathTracker} challenge}\label{sec:model_benchmark}
Can state-of-the-art models for video analysis match humans on \textit{PathTracker}? To test this question we surveyed a variety of architectures that are the basis for leading approaches to many video analysis tasks, from object tracking to action classification. We restricted our survey to models that could be trained end-to-end to solve \textit{PathTracker} without any additional pre- or post-processing steps. The selected models fall into three groups: (\textit{i}) deep convolutional networks (CNNs), (\textit{ii}) transformers, and (\textit{iii}) recurrent neural networks (RNNs). The deep convolutional networks include a 3D ResNet (R3D~\cite{Tran2017-fg}), a space/time separated ResNet with ``2D-spatial + 1D-temporal'' convolutions (R(2+1)D~\cite{Tran2017-fg}), and a ResNet with 3D convolutions in early residual blocks and 2D convolutions in later blocks (MC3~\cite{Tran2017-fg}). We trained versions of these models with random weight initializations and weights pretrained on ImageNet. We included an R3D trained from scratch without any downsampling, in case the small size of \textit{PathTracker} videos caused learning problems (see SI \S\ref{si_sec:challenge} for details). We also trained a version of the R3D on optic flow encodings of \textit{PathTracker} (SI \S\ref{si_sec:challenge}). For transformers, we turned to the TimeSformer~\cite{Bertasius2021-hi}. We test two of its instances: (\textit{i}) attention is jointly computed for all locations across space and time in videos, and (\textit{ii})  temporal attention is applied before spatial attention, which results in massive computational savings. We found similar \textit{PathTracker} performance with both models. We report the latter version here as it was marginally better (see SI \S\ref{si_sec:challenge} for performance of the other, joint space-time attention TimeSformer). We include a version of the TimeSformer trained from scratch, and a version pre-trained on ImageNet-20K. Note that state-of-the-art transformers for object tracking in natural videos feature similar deep and multi-headed designs~\cite{Wang2021-no} but use additional post-processing steps that are beyond the scope of \textit{PathTracker}. Finally, we include a convolutional-gated recurrent unit (Conv-GRU)~\cite{Bhat2020-hb}.  

\paragraph{Method} The visual simplicity of \textit{PathTracker} cuts two ways: it makes it possible to compare human and model strategies for tracking without re-recognition as long as the task is not too easy. Prior synthetic challenges like \textit{Pathfinder} constrain sample sizes for training to probe specific computations~\cite{Kim2020-yw,Linsley2018-ls,Tay2020-ni}. We adopt the following strategy to select a training set size that would help us test tracking strategies that do not depend on re-recognition. We took Inception 3D (I3D) networks~\cite{Carreira2017-ic}, which have been a strong baseline architecture in video analysis over the past several years, and tested their ability to learn \textit{PathTracker} as we adjusted the number of videos for training. As we discuss in SI \S\ref{si_sec:related_work}, when this model was trained with 20K examples of the 32 frame and 14 distractor version of \textit{PathTracker} it achieved good performance on the task without signs of overfitting to its simple visual statistics. We therefore train all models in subsequent experiments with 20K examples. 

We measure the ability of models to learn \textit{PathTracker} and systematically generalize to novel versions of the challenge when trained on 20K samples. We trained models using a similar approach as in our human psychophyics. Models were trained on one version of pathfinder, and tested on other versions with the same number of frames, and the same or different number of distractors. In the first experiment, models were trained on the 32 frame and 14 distractor \textit{PathTracker}, then tested on the 32 frame \textit{PathTracker} datasets with 1, 14, or 25 distractors (Fig.~\ref{fig:model_perf}a). In the second experiment, models were trained on the 64 frame and 14 distractor \textit{PathTracker}, then tested on the 64 frame \textit{PathTracker} datasets with 1, 14, or 25 distractors (Fig.~\ref{fig:model_perf}a). Models were trained to detect if the target dot reached the blue goal marker using binary crossentropy and the Adam optimizer~\cite{Kingma2014-ct} until performance on a test set of 20K videos with 14 distractors decreased for 200 straight epochs. In each experiment, we selected model weights that performed best on the 14 distractor dataset. Models were retrained three times on learning rates $\in \{1$e-$2, 1$e-$3, 1$e-$4, 3$e-$4, 1$e-$5\}$ to optimize performance. The best performing model was then tested on the remaining 1 and 25 distractor datasets in the experiment. We used four NVIDIA GTX GPUs and a batch size 180 for training.

\paragraph{Results} We treat human performance as the benchmark for models on \textit{PathTracker}. Nearly all CNNs and the ImageNet-initialized TimeSformer performed well enough to reach the 95\% human confidence interval on the 32 frame and 14 distractor \textit{PathTracker}. However, all models performed worse when systematically generalizing to \textit{PathTracker} datasets with a different number of distractors, even when that number decreased (Fig.~\ref{fig:model_perf}a, 1 distractor). Model performance on the 32 frame \textit{PathTracker} datasets was worst for 25 distractors. No CNN or transformer reached the 95\% confidence interval of humans on this dataset (Fig.~\ref{fig:model_perf}a). The optic flow R3D and the TimeSformer trained from scratch were even less successful but still above chance, while the Conv-GRU performed at chance. Model performance plummeted across the board on 64 frame \textit{PathTracker} datasets. 
The drop in model performance from 32 to 64 frames reflects a combination of the following features of \textit{PathTracker}. (\textit{i}) The target becomes more likely to cross a distractor when length and the number of distractors increase (Fig.~\ref{fig:dataset}; Fig.~2c). This makes the task difficult because the target is momentarily impossible to distinguish from a distractor. (\textit{ii}) The target object must be tracked from start-to-end to solve the task, which can incur a memory cost that is monotonic \wrt video length. (\textit{iii}) The prior two features interact to non-linearly increase task difficulty (Fig.~2c).



\begin{figure}[t]
\begin{center}\small
\includegraphics[width=.99\textwidth]{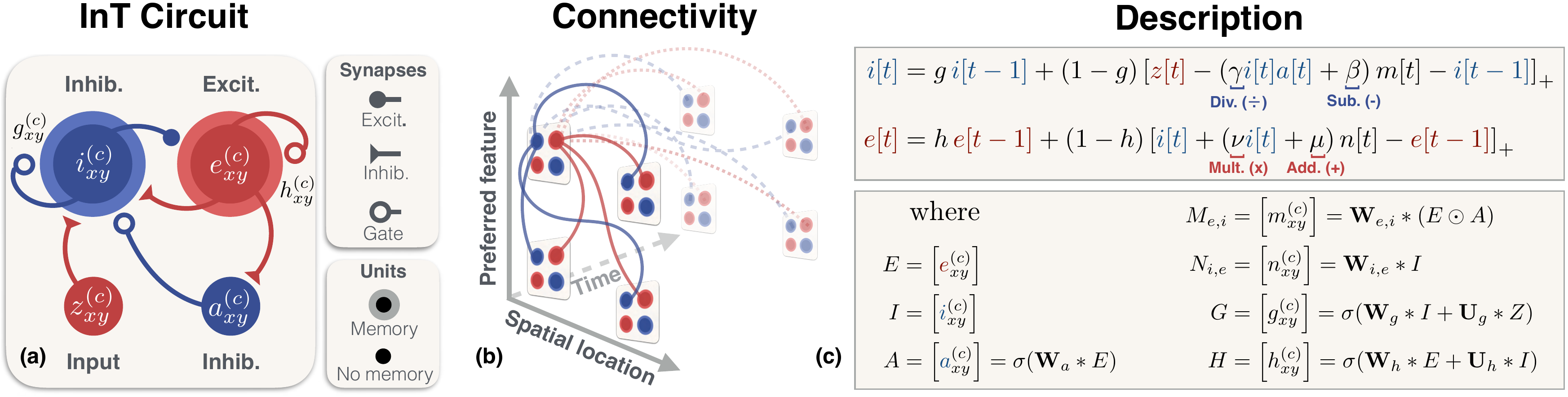}
\end{center}
\vspace{-4mm}
\caption{The Index-and-Track (InT) circuit model is inspired by Neuroscience models of motion perception~\cite{Berzhanskaya2007-lu} and executive cognitive function~\cite{Wong2006-xa}. (\textit{a}) The circuit receives input encodings from a video ($z$), which are processed by interacting recurrent inhibitory and excitatory units ($i, e$), and a mechanism for selective ``attention'' ($a$) that tracks the target location. (\textit{b}) InT units have spatiotemporal receptive fields. Spatial connections are formed by convolution with weight kernels ($W_e,W_i$). Temporal connections are controlled by gates ($g, h$). (\textit{c}) Model parameters are fit with gradient descent. Softplus$ = [.]$, sigmoid$ = \sigma$, convolution$ = *$, elementwise product $= \odot$.}\vspace{-4mm}
\label{fig:model}
\end{figure}

\paragraph{Neural circuits for tracking without re-recognition} \textit{PathTracker} is inspired by object tracking paradigms from Psychology, which tested theories of working memory and attention in human observers~\cite{Pylyshyn1988-pi,Blaser2000-xz}. \textit{PathTracker} may draw upon similar mechanisms of visual cognition in humans. However, the video analysis models that we include in our benchmark (Fig.~\ref{fig:model_perf}) do not have inductive biases for working memory, and while the TimeSformer uses a form of attention, it is insufficient for learning \textit{PathTracker} and only reached human performance on one version of the challenge (Fig.~\ref{fig:model_perf}).

Neural circuits for motion perception, working memory, and attention have been the subject of intense study in Neuroscience for decades. Knowledge synthesized from several computational, electrophysiological and imaging studies point to canonical features and computations that are carried out by these circuits. (\textit{i}) Spatiotemporal feature selectivity emerges from non-linear and time-delayed interactions between neuronal subpopulations~\cite{Takemura2013-ch,Kim2014-bc}. (\textit{ii}) Recurrently connected neuronal clusters can maintain task information in working memory~\cite{Elman1990-hd,Wong2006-xa}. (\textit{iii}) Synaptic gating, inhibitory modulation, and disinhibitory circuits are neural substrates of working memory and attention~\cite{Hochreiter1997-gc,OReilly2006-je,Badre2012-hv,DArdenne2012-az,Mitchell2007-fd}. (\textit{iv}) Mechanisms for gain control may aid motion-based object tracking by building tolerance to visual nuisances, such as illumination~\cite{Berzhanskaya2007-hs,Mely2018-bc}. We draw from these principles to construct the ``Index-and-Track'' circuit (InT, Fig.~\ref{fig:model}). 

\begin{figure}[t]
\begin{center}\small
\includegraphics[width=.99\textwidth]{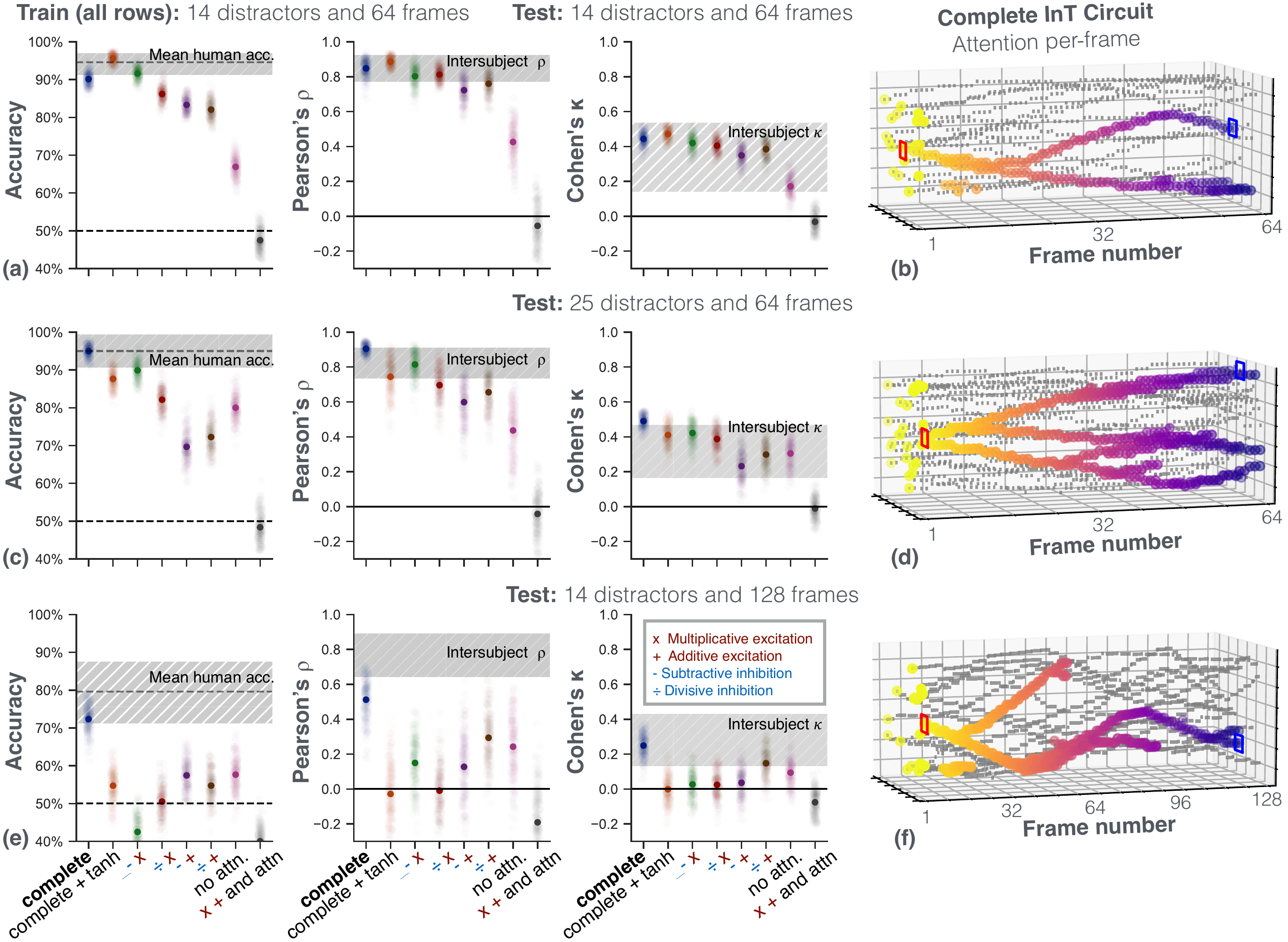}
\end{center}
\vspace{-4mm}
\caption{Performance, decision correlations, and error consistency between models and humans on \textit{PathTracker}. In a new set of psychopysics experiments, humans and models were trained on 64 frame \textit{PathTracker} datasets with 14 distractors, and rendered decisions on a variety of challenging versions. Decision correlations are computed with Pearson's $\rho$, and error consistency with Cohen's $\kappa$~\cite{Geirhos2020-uq}. Only the Complete InT circuit rivals human performance and explains the majority of their decision and error variance on each test dataset (\textit{a,c,e}). Visualizing InT attention ($a$) reveals that it has learned to solve \textit{PathTracker} by multi-object tracking (\textit{b,d,f}; color denotes time). The consistency between InT and human decisions raises the possibility that humans rely on a similar strategy.}\vspace{-4mm}
\label{fig:correlations}
\end{figure}

\vspace{-2mm}
\paragraph{InT circuit description} The InT circuit takes an input $z$ at location $x,y$ and feature channel $c$ from video frame $t \in T$ (Fig.~\ref{fig:model}a). This input is passed to an inhibitory unit $i$, which interacts with an excitatory unit $e$, both of which have persistent states that store memories with the help of gates $g,h$. The inhibitory unit is also gated by another inhibitory unit, $a$, which is a non-linear function of $e$, and can either decrease or increase (\ie through disinhibition) the inhibitory drive. In principle, the sigmoidal nonlinearity of $a$ means that it can selectively attend, and hence, we refer to $a$ as ``attention''.  Moreover, since $a$ is a function of $e$, which lags in time behind $z[t]$, its activity reflects the displacement (or motion) of an object in $z[t]$ versus the current memory of $e$. InT units have spatiotemporal receptive fields (Fig.~\ref{fig:model}b). Interactions between units at different locations are computed by convolution with weight kernels $\textbf{W}_{e,i},\textbf{W}_{i,e} \in \mathbb{R}^{5,5,c,c}$ and attention is computed by $\textbf{W}_a \in \mathbb{R}^{1,1,c,c}$. Gate activities that control InT dynamics and temporal receptive fields are similarly calculated by kernels, $\textbf{W}_{g},\textbf{W}_{h},\textbf{U}_{g},\textbf{U}_{h} \in \mathbb{R}^{1,1,c,c}$. Recurrent units in the InT support non-linear (gain) control. Inhibitory units can perform divisive and subtractive computations, controlled by $\gamma,\beta$. Excitatory units can perform multiplicative and additive computations, controlled by $\nu,\mu$. Parameters $\gamma,\beta,\nu,\mu \in \mathbb{R}^{c}$. ``SoftPlus'' rectifications denoted by $[.]_+$ enforce inhibitory and excitatory function and competition (Fig.~\ref{fig:model}c). The final $e$ state is passed to a readout for \textit{PathTracker} (SI \S\ref{si_sec:int}). 


\vspace{-2mm}
\paragraph{InT \textit{PathTracker} performance} We trained the InT on \textit{PathTracker} following the procedure in \S\ref{sec:model_benchmark}. It was the only model that rivaled humans on each version of \textit{PathTracker} (Fig.~\ref{fig:model_perf}). The gap in performance between InT and the field is greatest on the 64 frame version of the challenge.

How does the InT solve \textit{PathTracker}? There are at least two strategies that it could choose from. One is to maintain a perfect track of the target throughout its trajectory, and extrapolate the momentum of its motion to resolve crossings with distractors. Another is to track all objects that cross the target and check if any of them reach the goal marker by the end of the video. To investigate the type of strategy learned by the InT for \textit{PathTracker} and to compare this strategy to humans, we ran additional psychophysics with a new group of 90 participants using the same setup detailed in \S\ref{sec:challenge}. Participants were trained on 8 videos from the 14 distractor and 64 frame \textit{PathTracker} and tested on 72 images from either the (\textit{i}) 14 distractor and 64 frame dataset, (\textit{ii}) 25 distractor and 64 frame dataset, or (\textit{iii}) 14 distractor and 128 frame dataset. Unlike the psychophysics in \S\ref{sec:challenge}, all participants viewing a given test set saw the same videos, which made it possible to compare their decision strategies with the InT. 

InT performance reached the $95\%$ confidence intervals of humans on each test dataset. The InT also produced errors that were extremely consistent 
with humans and explained nearly all variance in Pearson's $\rho$ and Cohen's $\kappa$ on each dataset (Fig.~\ref{fig:correlations}, middle and right columns). This result means that humans and InT rely on similar strategies for solving \textit{PathTracker}. 

What is the underlying strategy? We visualized activity of $A$ units in the InT as they processed \textit{PathTracker} videos and found that they had learned a multi-object tracking strategy to solve the task (Fig.~\ref{fig:correlations}; see SI \S\ref{si_sec:attention} for the method, and \url{http://bit.ly/InTcircuit} for animations). The $A$ units track the target object until it crosses a distractor and ambiguity arises, at which point attention splits and it tracks both objects. This strategy indexes a limited number of objects at once, consistent with studies of object tracking in humans~\cite{Pylyshyn1988-pi}. Since the InT is not explicitly constrained for this tracking strategy, we next investigated the minimal circuit for learning it and explaining human behavior. 


We developed versions of the InT with lesions applied to different combinations of its divisive/subtractive and multiplicative/additive computations, a version without attention units $A$, and a version that does not make a distinction of inhibition \vs excitation (``complete + tanh''), in which rectifications were replaced with hyperbolic tangents that squash activities into $[-1, 1]$. While some of these models marginally outperformed the Complete InT on the 14 distractor and 64 frame dataset, their performance dropped precipitously on the 25 distractor and 64 frame dataset, and especially the very long 14 distractor and 128 frame dataset (Fig.~\ref{fig:correlations}e). Attention units in the complete InT's nearest rival (complete + tanh) were non-selective, potentially contributing to its struggles. InT performance also dropped when we forced it to attend to fewer objects (SI \S\ref{si_sec:tracking}).


\begin{figure}[t]
\begin{center}\small
\includegraphics[width=.99\textwidth]{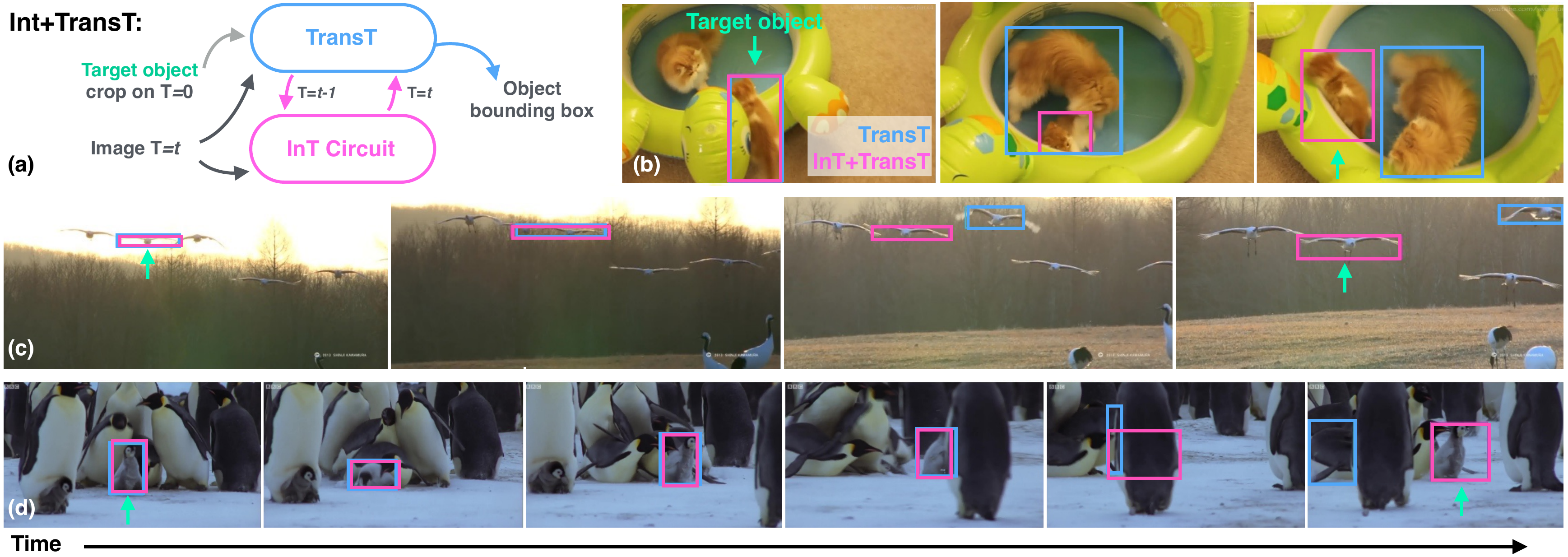}
\end{center}
\vspace{-4mm}
\caption{Circuit mechanisms for tracking without re-recognition build tolerance to visual nuissances that affect object appearance. (\textit{a}) The TransT\cite{Chen2021-is} is a transformer architecture for object tracking. We develop an extension, the InT+TransT, in which our InT circuit model recurrently modulates TransT activity. Unlike the TransT, the InT+TransT is trained on sequences to promote tracking strategies that do not rely on re-recognition. (\textit{b-d}) The InT+TransT excels when the target object is visually similar to other depicted objects, undergoes changes in illumination, or is occluded.}\vspace{-4mm}
\label{fig:got}
\end{figure}

\section{Appearance-free mechanisms for object tracking in the wild}\label{sec:wild}
The InT solves \textit{PathTracker} by learning to track multiple objects at once, without relying on the re-recognition strategy that has been central to progress in video analysis challenges in computer vision. However, it is not clear if tracking without re-recognition is useful in the natural world. We test this question by turning to object tracking in natural videos. At the time of writing, the state-of-the-art object tracker is the TransT~\cite{Chen2021-is}, a deep multihead transformer~\cite{Vaswani2017-gh}. The TransT finds pixels in a video frame that match the appearance of an image crop depicting a target object. During training, the TransT receives a tuple of inputs, consisting of this target object image and a random additional ``search frame'' from the same video. These images are encoded with a modified ResNet50~\cite{He2015-cb}, passed to separate transformers, and finally combined by a ``cross-feature attention'' (CFA) module, which compares the two encodings via a transformer key/query/value computation. The target frame is used for key and value operations, and the search frame is used for the query operation. Through its pure appearance-based approach to tracking, the TransT has achieved top performance on
TrackingNet~\cite{Muller2018-qn}, LaSOT~\cite{Fan2018-sb}, and GOT-10K~\cite{Huang2018-yt}.

\paragraph{InT+TransT} We tested whether or not the InT circuit can improve TransT performance by learning a complementary object strategy that does not depend on appearance, or re-recognition. We reasoned that this strategy might help TransT tracking in cases where objects are difficult to discern by their appearance, such as when they are subject to changing lighting, color, or occlusion. We thus developed the InT+TransT, which involves the following modifications of the original TransT (Fig.~\ref{fig:got}a). (\textit{i}) We introduce two InT circuits to form a bottom-up and top-down feedback loop with the TransT~\cite{Linsley2020-en,Gilbert2013-hb}, which in principle will help the model select the appropriate tracking strategy depending on the video -- re-recognition or not. One InT receives ResNet50 search image encodings and modulates the TransT's CFA encoding of this search image. The other receives the output of the TransT and uses this information to update memory in the first InT. (\textit{ii}) The TransT is trained with pairs of target and search video frames, separated in time by up to 100 frames. We introduce the intervening frames to the InT circuits. See SI \S\ref{si_sec:tracking} for extended methods.

\paragraph{Training} InT+TransT training and evaluation hews close to the TransT procedure. This includes training on the latest object tracking challenges in computer vision: TrackingNet~\cite{Muller2018-qn}, LaSOT~\cite{Fan2018-sb}, and GOT-10K~\cite{Huang2018-yt}. All three challenges depict diverse classes of objects, moving in natural scenes that range from simplistic and barren to complex and cluttered. TrackingNet (30,132 train and 511 test videos) and GOT-10K (10,000 train and 180 test) evaluation is performed on official challenge servers, whereas LaSOT (1,120 train and 280 test) is evaluated with a Matlab toolbox. While the TransT is also trained with static images from Microsoft COCO~\cite{Lin2014-zk}, in which the search image is an augmented version of the target, we do not include COCO in InT+TransT since we expect object motion to be an essential feature for our model~\cite{Chen2021-is}. The InT+TransT is initialized with TransT weights and trained with AdamW~\cite{Loshchilov2017-dy} and a learning rate of 1$e-$4 for InT parameters, and 1$e-$6 for parameters in the TransT readout and CFA module. Other TransT parameters are frozen and not trained. The InT+TransT is trained with the same objective functions as the TransT for target object bounding box prediction in the search frame, and an additional objective function for bounding box prediction using InT circuit activity in intervening frames. The complete model was trained with batches of 24 videos on 8 NVIDIA GTX GPUs for 150 epochs (2 days). We selected the weights that performed best on GOT-10K validation. A hyperparameter controls the number of frames between the target and search that are introduced into the InT during training. We relied on coarse sampling (1 or 8 frames) due to memory issues associated with recurrent network training on long sequences~\cite{Linsley2020-ua}.

\paragraph{Results} An InT+TransT trained on sequences of 8 frames performed inference around 30FPS on a single NVIDIA GTX and beat the TransT on nearly all benchmarks. It is in first place on the TrackingNet leaderboard (\url{http://eval.tracking-net.org/}), better than the TransT on LaSOT, and rivals the TransT on the GOT-10K challenge (Table~\ref{table:tracking}). The InT+TransT performed better when trained with longer sequences (compare $T=8$ and $T=1$, Table~\ref{table:tracking}). Consistent with InT success on \textit{PathTracker}, the InT+TransT was qualitatively better than the TransT on challenging videos where the target interacted with other similar looking objects (Fig.~\ref{fig:got}; \url{http://bit.ly/InTcircuit}).

We also found that the InT+TransT excelled in other challenging tracking conditions. The LaSOT challenge provides annotations for challenging video features, which reveal that the InT+TransT is especially effective for tracking objects with ``deformable'' parts, such as moving wings or tails (SI \S\ref{si_sec:tracking}). We further test if introducing object appearance perturbations to the GOT-10K might distinguish performance between the TransT and InT+TransT. We evaluate these models on the GOT-10K test set with one of three perturbations: inverting the color of all search frames (Color), inverting the color of random search frames (rColor), or introducing random occlusions (Occl.). The InT+TransT outperformed the TransT on each of these tests (Table~\ref{table:tracking}).

\begin{table}
\centering
\begin{tabular}{l|llllll}\label{table:tracking}
\textbf{Model}  & \textbf{TrackingNet}\cite{Muller2018-qn} & \textbf{LaSOT}\cite{Fan2018-sb} & \textbf{GOT}\cite{Huang2018-yt} & $\substack{\mathrm{\mathbf{\color{white}{GOT}}}\\\mathrm{Color}\mathstrut}$ & $\substack{\mathrm{\textbf{GOT}}\\\mathrm{rColor}\mathstrut}$ & $\substack{\mathrm{\mathbf{\color{white}{GOT}}}\\\mathrm{Occl.}\mathstrut}$  \\ 
\hline
InT+TransT$_{T=8}$ & \textbf{87.5}                & 74.0           & \color{gray}{72.2}            & 43.1              & 62.5                     & 56.9                   \\
InT+TransT$_{T=1}$ & \color{gray}{87.3}                & \color{gray}{73.6}             & \color{gray}{70.0}              & \color{gray}{36.2}                  & \color{gray}{37.8}                         & \color{gray}{25.4}                       \\
TransT~\cite{Chen2021-is}          & \color{gray}{86.7}                & \color{gray}{73.8}             & 72.3            & \color{gray}{40.7}              & \color{gray}{57.5}                     & \color{gray}{55.2}                  
\end{tabular}
\caption{Model performance on TrackingNet (P$_{norm}$), LaSot (P$_{norm}$), GOT-10K (AO), and perturbations applied to the GOT-10K (AO). Best performance is in black, and certified state of the art is bolded. Perturbations on the GOT-10K are color inversions on every frame ($Color$) or random frames ($rColor$), and random occluders created from scrambling image pixels ($Occl.$). InT+TransT$_{T=8}$ was trained on sequences of 8 frames, and InT+TransT$_{T=1}$ was trained on 1-frame sequences.}\vspace{-4mm}
\end{table}
\vspace{-2mm}
\section{Discussion}\vspace{-2mm}
A key inspiration for our study is the centrality of visual motion and tracking across a broad phylogenetic range, via three premises: (\textit{i}) Object motion integration over time \textit{per se} is essential for ecological vision and survival~\cite{Lettvin1959-ha}. (\textit{ii}) Object motion perception cannot be completely reduced to recognizing similar appearance features at two different moments in time. In perceptual phenomena like \textit{phi} motion, the object that is tracked is described as ``formless'' with no distinct appearance~\cite{Steinman2000-jt}. (\textit{iii}) Motion integration over space and time is a basic operation of neural circuits in biological brains, which can be independent of appearance~\cite{Huk2005-bc}. These three premises form the basis for our work. 

We developed \textit{PathTracker} to test whether state-of-the-art models for video analysis can solve a visual task when object appearance is ambiguous. Prior visual reasoning challenges like \textit{Pathfinder}~\cite{Linsley2018-ls,Kim2020-yw,Tay2020-ni}, indicate that this is a problem for object recognition models, which further serve as a backbone for many video analysis models. While no existing model was able to contend with humans on \textit{Pathfinder}, our InT circuit was. Through lesioning experiments, we discovered that the InT's ability to explain human behavior depends on its full array of inductive biases, helping it learn a visual strategy that indexes and tracks a limited number of the objects at once, echoing classic theories on the role of attention and working memory in object tracking~\cite{Blaser2000-xz,Pylyshyn1988-pi}.


We further demonstrate that the capacity for video analysis without relying on re-recognition helps in natural scenes. Our InT+TransT model is more capable than the TransT at tracking objects when their appearance changes, and is the state of the art on the TrackingNet challenge. Together, our findings demonstrate that object appearance is a necessary element for for video analysis, but it is not sufficient for modeling biological vision and rivaling human performance. 

\begin{ack}
We are grateful to Daniel Bear for his suggestions to improve this work. We would also like to thank Rajan Girsa for initial discussions related to Python Flask used in MTurk portal. GM is also affiliated with Labrynthe Pvt. Ltd., New Delhi, India. Funding provided by ONR grant \#N00014-19-1-2029, the ANR-3IA Artificial and Natural Intelligence Toulouse Institute, and ANITI (ANR-19-PI3A-0004). Additional support from the Brown University Carney Institute for Brain Science, Center for Computation in Brain and Mind, and Center for Computation and Visualization (CCV).
\end{ack}

{\small
\bibliographystyle{splncs}
\bibliography{egbib}
}

\clearpage
\setcounter{figure}{0}
\makeatletter 
\renewcommand{\thefigure}{S\@arabic\c@figure}
\makeatother
\setcounter{table}{0}
\makeatletter 
\renewcommand{\thetable}{S\@arabic\c@table}
\renewcommand{\thefigure}{S\arabic{figure}}
\makeatother

\appendix

\section{Extended related work}\label{si_sec:related_work}

\paragraph{Translating circuits for biological vision into artificial neural networks} While the \textit{Pathfinder} challenge of~\cite{Linsley2018-ls} presents immense challenges for transformers and deep convolutional networks~\cite{Kim2020-yw}, the authors found that it can be solved by a simple model of intrinsic connectivity in visual cortex, with orders-of-magnitude fewer parameters than standard models for image categorization. This model was developed by translating descriptive models of neural mechanisms from Neuroscience into an architecture that can be fit to data using gradient descent~\cite{Linsley2018-ls,Linsley2020-ua}. Others have found success in modeling object tracking by drawing inspiration from ``dual stream'' theories of appearance and motion processing in visual cortex~\cite{Kosiorek2017-rb,Simonyan2014-nn}, or basing the architecture off of a partial connectome of the drosophila visual system~\cite{Tschopp2018-mo}. We adopt a similar approach in the current work, identifying mechanisms for object tracking without re-recognition in Neuroscience, and developing those into differentiable operations with parameters that can be optimized by gradient descent. This approach has the dual purpose of introducing task-relevant inductive biases into computer vision models, and developing theory on their relative utility for biological vision. 

\paragraph{Multi-object tracking in computer vision} The classic psychological paradigms of multi-object tracking~\cite{Pylyshyn1988-pi} motivated the application of models, like Kalman filters, which had tolerance to object occlusion when they relied on momentum models~\cite{Fortmann1980-yd}. However, these models are computationally expensive, hand-tuned, and because of this, not commonly used in computer vision anymore~\cite{Milan2016-cf}. More recent approaches include flow tracking on graphs~\cite{Zhang2008-ds} and motion tracking models that are relatively computationally efficient~\cite{Dicle2013-co}. However, even current approaches to multi-object tracking are not learned, instead relying on extensive hand tuning~\cite{Luo2021-ll}. In contrast, the point of \textit{PathTracker} is to understand the extent to which state-of-the-art neural networks are capable of tracking a single object in an array of distractors. Moreover, the solution discovered by the InT distinctly contrasts with these multi-object tracking models: it learns to index and track potential targets, rather than rely on momentum or other features that multi-object tracking models hard code. This is especially notable, since the InT is a model of neural circuit mechanisms, and its learned strategy predicts how humans solve the same task, and what neural systems are involved.

\paragraph{Limitations} In this work we tested a relatively small number of \textit{PathTracker} versions. We mostly focused on small variations to the number of distractors and video length, but in future work we hope to incorporate other variations like speed and velocity manipulations, and generalization across temporal variations. Another limitation is that appearance-free strategies confer relatively modest gains over the state of the art. One potential issue is determining when a visual system should rely on appearance-based \vs appearance-free features for tracking. Our solution is two-pronged and potentially insufficient. The first strategy is for top-down feedback from the TransT into the InT, which we aligns tracks between the two models. The second strategy is potentially naive, in that we gate the InT modulation to the TransT based on its agreement with the prior TransT query, and the confidence of the TransT query. Additional work is needed to identify better approaches. Meta-cognition work from Cognitive Neuroscience is one possible resource~\cite{Shimamura2000-ii}.

\paragraph{Societal impacts} The basic goal of our study is for understanding how biological brains work. \textit{PathTracker} helps us screen models against humans on a simple visual task which tests visual strategies for tracking without ``re-recognition'', or appearance cues. The fact that we developed a circuit that explains human performance is primarily important because it makes predictions about the types of neural circuit mechanisms that we might ultimately find in the brain in future Neuroscience work. Our extension to natural videos achieves new state-of-the-art because it is able to implement visual strategies that build tolerance to visual nuisances in way that resembles humans. It must be recognized the further development of this model has potential for misuse. One possible nefarious application is for surveillance. On the other hand, such a technology could be essential for ecology, sports, self-driving cars, robotics, and other real-world applications of machine vision. We open source our code and data to promote research towards such beneficial applications.

\begin{figure}[t]
\begin{center}\small
\includegraphics[width=.99\textwidth]{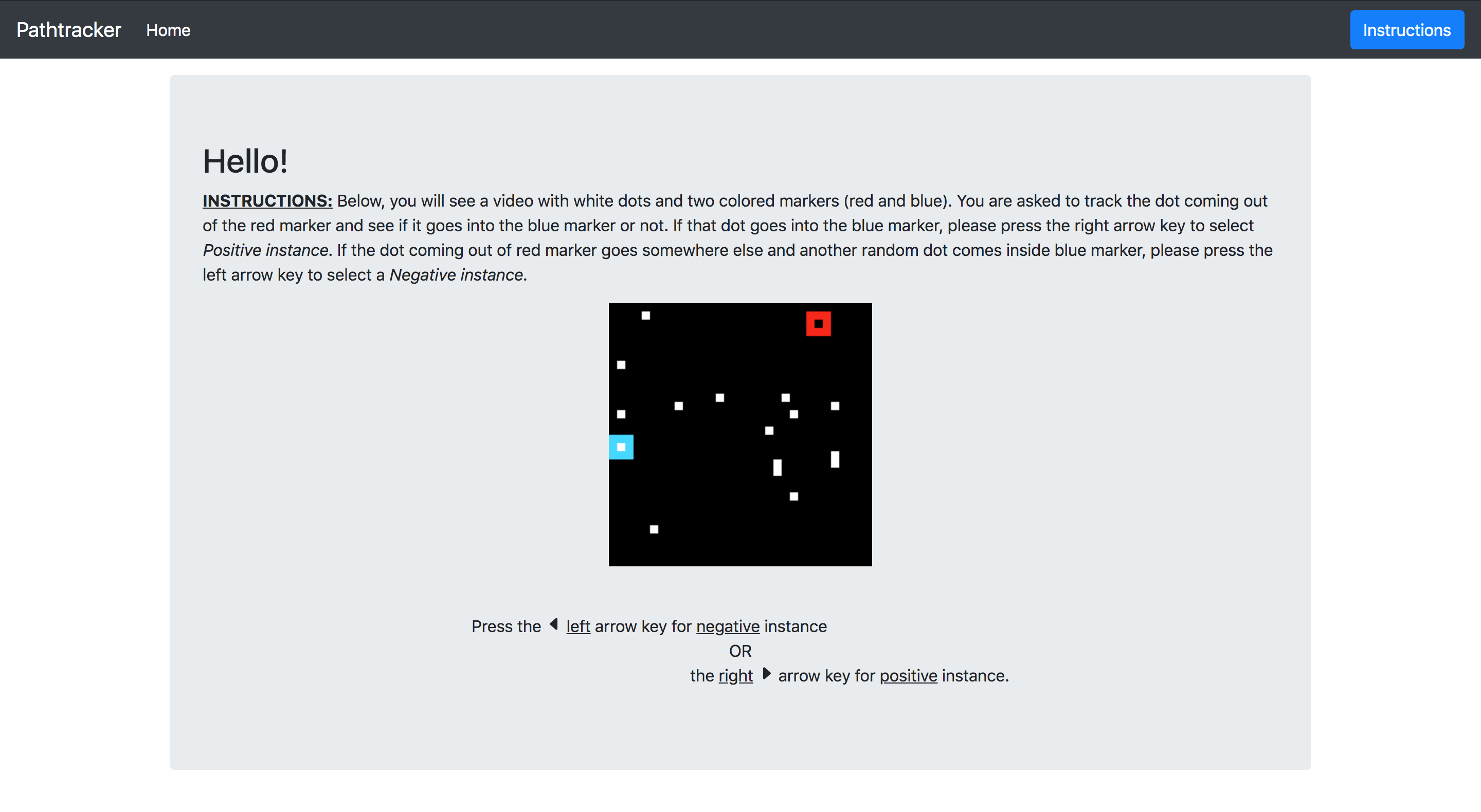}
\end{center}
\vspace{-4mm}
\caption{An experimental trial screen.}\vspace{-4mm}
\label{si_fig:portal_main}
\end{figure}

\section{Human benchmark}\label{si_sec:benchmark}
For our benchmark experiments we recruited 120 participants. Every participant was compensated with \$8 through MTurk on successful completion of all test trials by pasting a unique code generated by the system into their MTurk account. The decision regarding this amount was reached upon by prorating the minimum wage. An additional overhead fee of 40\% per participant was paid to MTurk. Collectively, we spent \$960 on these benchmark experiments.



The experiment was not time bound and participants could complete it at their own pace, taking around 25 minutes to complete. Videos with 32-, 64- and 128-frames were of duration 4, 8 and 14 seconds respectively. The videos played at 10 frames per second. Participant reaction times were also recorded on every trial and we include these in our data release. After every trial participants were redirected to a screen confirming successful submission of their response. They could start the next trial by clicking the ``Continue'' button or by pressing spacebar. If not, they were automatically redirected to the next trial after 3000 ms. Participants were also shown a ``rest screen'' with a progress bar after every 10 trials where they could take additional and longer breaks if needed. The timer was turned off for the rest screen. 

\paragraph{Experiment design} At the beginning of the experiment, we collected participant consent using a consent form approved by a University’s Institutional Review Board (IRB). Our experiment was completed on a computer via Chrome browser. Once consented, we provided a demonstration clearly stating the instructions with an example video to the participants. We also provided them with an option to revisit the instructions, if needed, from the top right corner of the navigation bar at any point during the experiment. 

Participants were asked to classify the video as ``positive'' (the dot leaving the red marker entered the blue marker) or ``negative'' (the dot leaving the red marker did not enter the blue marker) using the right and left arrow keys respectively. The choice for keys and their corresponding instances were mentioned below the video on every screen, along with a small instruction paragraph above the video. See Fig. \ref{si_fig:portal_main}. Participants were given feedback on their response (correct/incorrect) after every practice trial, but not after the test trials.

\paragraph{Setup} The experiment was written in Python Flask, including the server side script and logic. The frontend templates were written in HTML with Bootstrap CSS framework. We used javascript for form submission with keys and redirections, done on the end-user side. The server was run with nginx on 1 Intel(R) Xeon(R) CPU E5-2695 v3 at 2.30GHz, 4GB RAM, Red Hat Enterprise Linux Server. 

Video frames for each experiment were generated at 32$\times$32 resolution. Before writing them to the mp4 videos displayed to human participants in the experiment, the frames were resized through nearest-neighbor interpolation to 256$\times$256. In order to allow time for users to prepare for each trial, the first frame of each video was repeated 10 times before the rest of the video played.

\begin{figure}[t]
\begin{center}\small
\includegraphics[width=.99\textwidth]{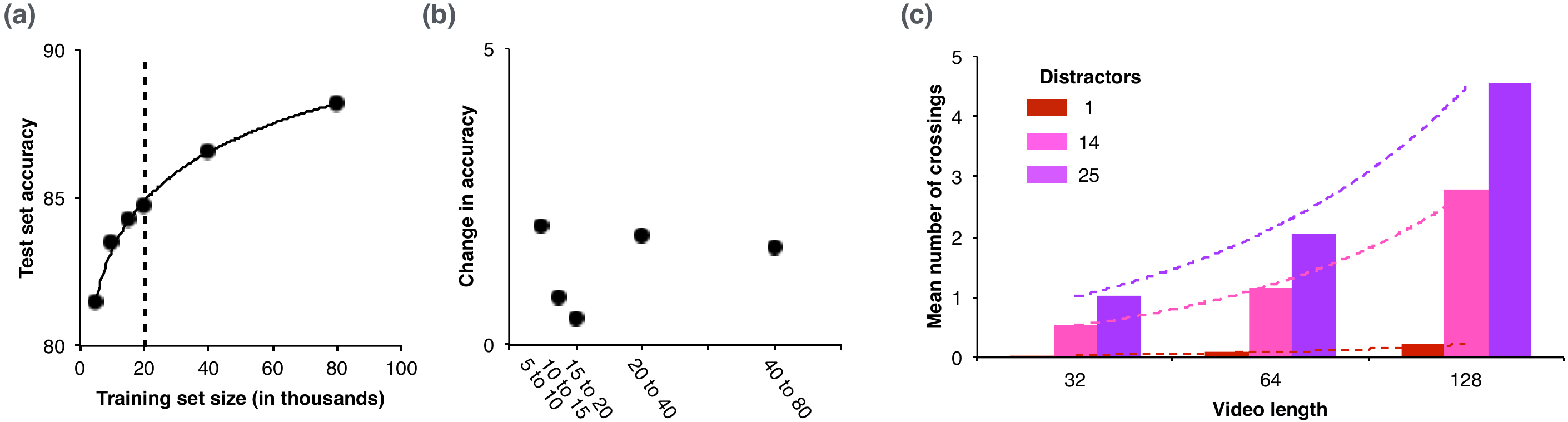}
\end{center}
\vspace{-4mm}
\caption{Our approach for selecting training set size on \textit{PathTracker}, and a proxy for difficulty across the versions of the challenge. (\textit{a}) We plot I3D performance as a function of training set size. The dotted line denotes the point at which the derivative of accuracy \wrt training set size is smallest (\textit{b}). We take this change performance as a function of training set size as evidence that I3D has learned a strategy that is sufficient for the task. We suspected this size would make the \textit{PathTracker} challenging but still solvable for the models we discuss in the main text. (\textit{c}) The number of average crossings in \textit{PathTracker} videos as a function of distractors and video length. Lines depict exponential fits for each number of distractors across lengths.}\vspace{-4mm}
\label{si_fig:extras}
\end{figure}

\paragraph{Filtering criteria} Amazon Mechanical Turk data is notoriously noisy. Because of this, we adopted a simple and bias-free approach to filter participants who were inattentive or did not understand the task (these users were still paid for their time). For the main benchmark described in \S3 in the main text, participants completed one of two experiments, where they were trained and tested on videos with 32 or 64 frames. No participant viewed both lengths of \textit{PathTracker}. Participants were trained with 14 distractor videos, then tested on videos with 1, 14, or 25 distractors. We filtered participants according to their performance on the \textit{training videos} for a particular experiment, which were otherwise not used for any analysis in this study. We removed participants who did not exceed 2 median absolute deviations below the median $median(X) - 2*MAD(X)$ (MAD $=$ median absolute deviation~\cite{Rousseeuw1993-wj}; this is a robust alternative to using the mean and standard deviation to find outliers). The threshold was approximately $40\%$ \textit{training} accuracy for each experiment (chance is $50\%$). This procedure filtered 74/180 participants in the benchmark.

\paragraph{Statistical testing} We assessed the difference between human performance and chance using randomization tests~\cite{Edgington1964-zb}. We computed human accuracy on each test dataset, then over 10,000 steps, we shuffled video labels, and then recomputed and stored the resulting accuracy. We computed $p-$values as the proportion of shuffled accuracies that exceed the real accuracy. We also used linear models for significance testing of trends in human accuracy as we increased the number of distractors. From these models we computed $t$-tests and $p$-values.

\paragraph{Using an I3D~\cite{Carreira2017-ic} to select \textit{PathTracker} training set sizes}
As mentioned in the main text, we selected \textit{PathTracker} training set size for models reported in the main text by investigating sample efficiency of the standard but not state-of-the-art I3D~\cite{Carreira2017-ic}. We were specifically interested in identifying a ``pareto principle'' in learning dynamics where additional training samples began to yield smaller gains in accuracy, potentially signifying a point at which I3D had learned a viable strategy (SI Fig.~\ref{si_fig:extras}). At this point, we suspected that the task would remain challenging -- but still solvable -- across the variety of \textit{PathTracker} conditions we discuss in the main text. We focus on basic 32 frame and 14 distractor training and find an inflection point at 20K examples. We plot I3D performance on this condition in SI Fig.~\ref{si_fig:extras}a and performance slopes in SI Fig.~\ref{si_fig:extras}b. The first and lowest slope corresponds to 20K samples, and hence may reflect an inflection in the model's visual strategy. Our experiments in the main text demonstrate that this strategy is a viable one for calibrating the difficulty of synthetic challenges.

\paragraph{Target-distractor crossings}
We compute the number of average crossings between the target object and distractors in \textit{PathTracker}. Increasing video length monotonically increases the number of crossings. Length further interacts with the number of distractors to yield more crossings (SI Fig.~\ref{si_fig:extras}c).

\section{Solving the Pathtracker challenge}\label{si_sec:challenge}

\begin{figure}[t]
\begin{center}\small
\includegraphics[width=.99\textwidth]{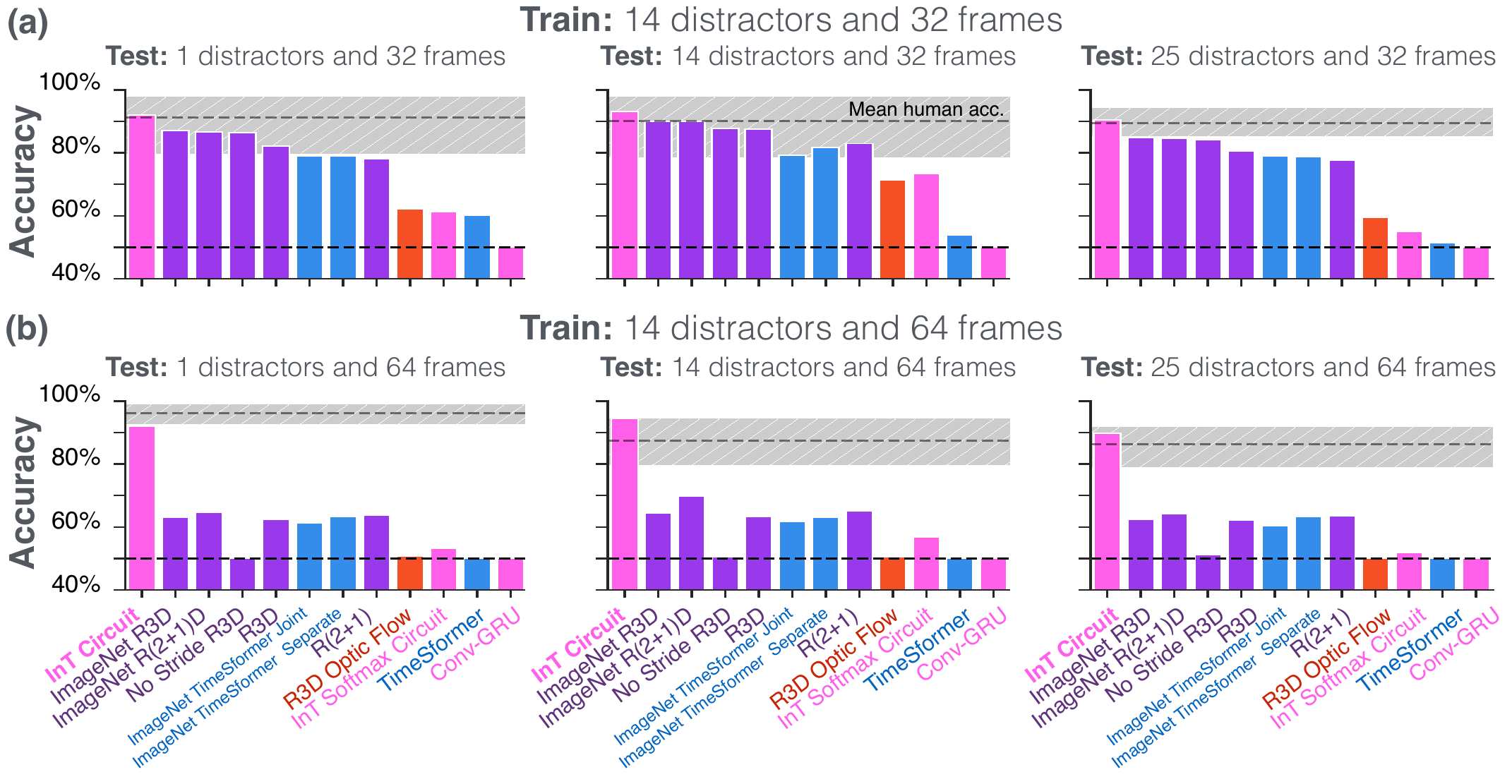}
\end{center}
\vspace{-4mm}
\caption{An extended benchmark of state-of-the-art models on \textit{PathTracker} with (\textit{a}) 32 and (\textit{b}) 64 frame versions of the task.}\vspace{-4mm}
\label{si_fig:extended_benchmark}
\end{figure}

\paragraph{State-of-the-art model details} We trained a variety of models on our benchmark. This included an R3D without any strides or downsampling. Because this manipulation caused an explosion in memory usage, we reduced the number of features per-residual block of this ``No Stride R3D'' from 64/128/256/512 to 32/32/32/32. We also included two forms of TimeSformers~\cite{Bertasius2021-hi}, one with distinct applications of temporal and spatial attention that we include in our main analyses, and another with join temporal and spatial attention (SI Fig.~\ref{si_fig:extended_benchmark}).

\paragraph{Optic Flow} We followed the method of \cite{Carreira2017-ic} to compute optic flow encodings of \textit{PathTracker} datasets. We used OpenCV's implementation of the TV-L1 algorithm \cite{Wedel2009-nq}. We extracted two channels from the output given by the algorithm, and appended a channel-averaged version of the corresponding \textit{PathTracker} image, similar to the approach of \cite{Carreira2017-ic}.

\begin{figure}[t]
\begin{center}\small
\includegraphics[width=.99\textwidth]{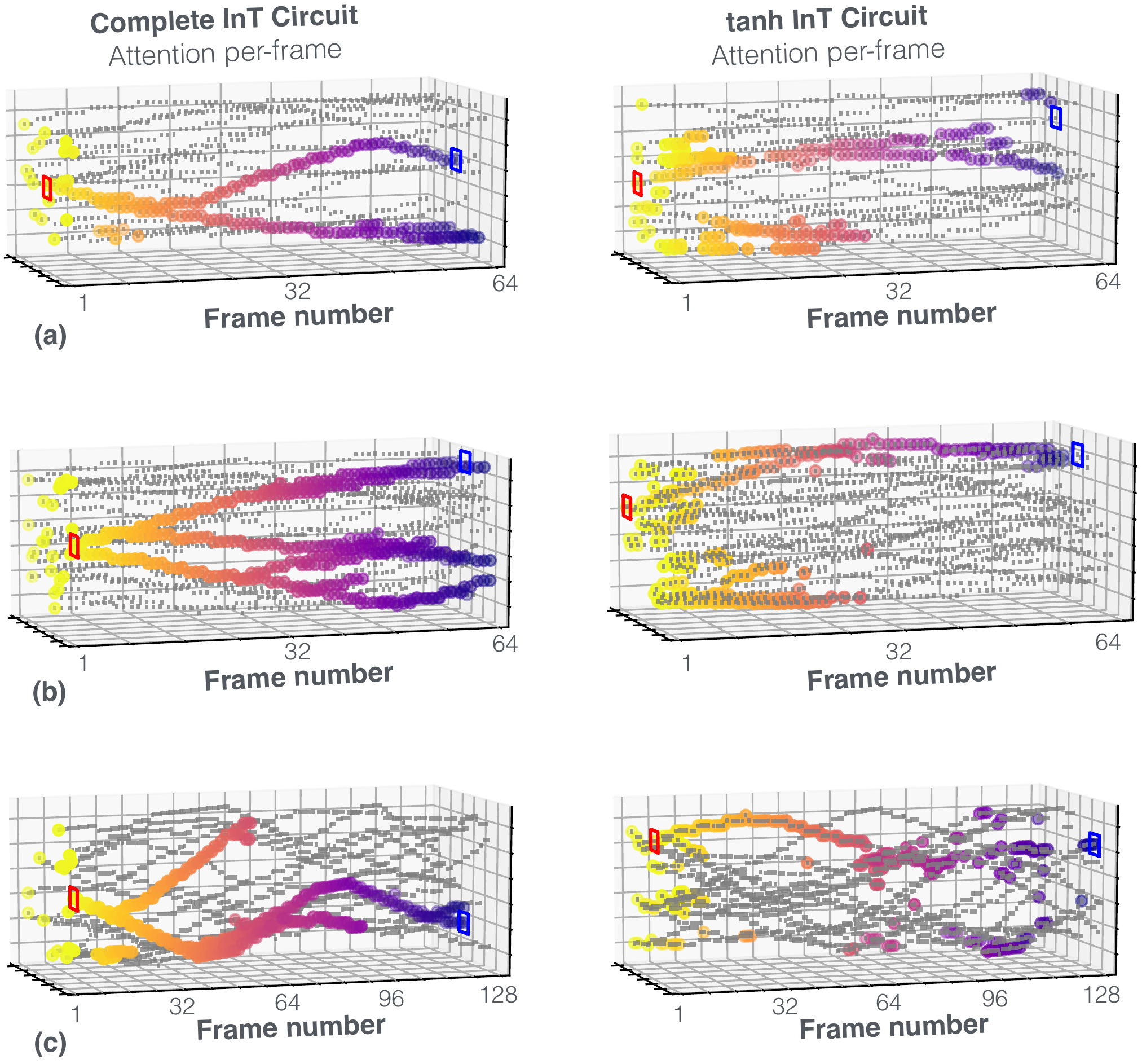}
\end{center}
\vspace{-4mm}
\caption{A comparison of attention between the complete InT and one where its softplus rectifications are replaced by tanh.}\vspace{-4mm}
\label{si_fig:attention}
\end{figure}

\section{InT circuit description}\label{si_sec:int}
Our InT circuit has two recurrent neural populations, $I$ and $E$. These populations evolve over time and receive a dynamic ``feedforward'' drive via $Z$. This feedforward drive is derived from a convolution between each frame of the \textit{PathTracker} videos a kernel $W_z \in \mathbb{R}^{1,1,3,32}$. This activity is then rectified by a softplus pointwise nonlinearity. InT hidden states are initialized with $0.6931=\mathrm{softplus}(0)$. The InT circuit also includes Batch Normalization~\cite{Ioffe2015-zm} applied to the outputs of its recurrent kernels $W_ie,W_ei$, with scales ($\alpha$) and intercepts ($\eta$) shared across timesteps of processing. We initialize the scale parameters to $0.1$ following prior work~\cite{Linsley2020-ua}. We do not store Batch Normalization moments during training. InT gain control (\ie its divisive normalization) is expected to emerge at steady state~\cite{Mely2018-bc,Grossberg1985-ui} in similar dynamical systems formulations, although our formulation relaxes some of these constraints.

The final activity of $E[T]$ in the InT for a \textit{PathTracker} video is passed to a readout that renders a binary decisions for the task. This readout begins by convolving $E[T]$ with a kernel $W_{r1} \in \mathbb{R}^{1,1,32,1}$. The output is channel-wise concatenated with the channel of the first frame containing the location of the goal marker. This activity is then convolved with another kernel $W_{r2} \in \mathbb{R}^{5,5,2,1}$, which is designed to capture overlap between the goal marker and the putative target object/dot. The resulting activity is ``global'' average pooled and entered into binary crossentropy for model optimization. On \textit{PathTracker}, all versions of the InT and the ConvGRU used this input transformation. All versions of the InT, the ConvGRU, and the ``No Stride R3D' used this readout. 

\paragraph{Spatiotemporal filtering through recurrent connections} An open question is whether recurrent neural networks with convolutional connections are capable of learning tuned spatiotemporal feature selectivity. That is, the ability to learn to detect a specific visual feature moving in a certain direction. Adelson and Bergen~\cite{Adelson1985-ot} laid out a plausible solution, in which spatial filters offset by phase are combined over time through positive or negative weights. The success of our InT on \textit{PathTracker} indicates that it might have adopted a similar solution, using its horizontal connection kernels $W_{i,e},W_{e,i}$ to learn spatial filters offset in phase (\eg an on-center off-surround and an off-center on-surround filter), which are combined via learned gates to yield spatiotemporal tuning. We leave an analysis of the InT ``connectome'' as it relates to spatiotemporal feature learning to future work.

\paragraph{Deriving the InT} 
For the sake of clarity and succinctness, we focus the derivation of the $InT$ circuit's update equations to reflect that of generic single neurons, which without loss of generality applies to the each spatial/feature dimension. The $InT$ circuit model is built on top of two recurrent populations ($E/I$) of neurons (serving excitatory/ inhibitory roles respectively), and a state-less population of neurons ($A$) that serves as an attentional controller. We denote these populations as follows:
\begin{align}
E = \lb e_{xy}^{(c)} \rb; 
I = \lb i_{xy}^{(c)} \rb; 
A = \lb a_{xy}^{(c)} \rb
\end{align}

Here, the $x,y$ subscripts denote spatial tuning, and the $c$ superscript denotes feature tuning. Moving forward, we reference generic units from these populations with $e$, $i$, and $a$ respectively. In essence, the circuit can be expressed as a continuous first-order coupled differential system of this form.

\begin{align}
\begin{split}
\tau_{inh} \frac{di}{dt}
&= -i + \lb z - (\gamma i a + \beta)m \rb_+ 
\\
\tau_{exc} \frac{de}{dt} &= -e + \lb i + (\nu i + \mu)n \rb_+
\label{si_eq:continoustime}
\end{split}
\end{align}

In Eq.~\ref{si_eq:continoustime}, $\gamma$, $\beta$, $\nu$, and $\mu$ are model hyperparameters, while $m$ and $n$ are themselves functions of $e$, $i$, and $a$. The exact functional form of $m$ and $n$ is detailed in Fig.~\ref{fig:model}b in the main text. $z$ is an external input to the system.

For the purposes of simulating and training this model with gradient descent, we use a first-order Euler approximation with time step $\Delta t$. Assuming we choose $g = \frac{\Delta t}{\tau_{inb}}$ and $h = \frac{\Delta t}{\tau_{exc}}$, the discretized version of Eq.~\ref{si_eq:continoustime} can be written as follows.

\begin{align}
\begin{split}
i_t
&= (1 - g) \, i_{t-1} + g \lb z_t - (\gamma i_t a_t + \beta)\, m_t \rb_+ \
\\
e_t &= (1 - h) \, e_{t-1} + h \lb i_t + (\nu i_t + \mu)\, n_t \rb_+
\label{si_eq:discretetime}
\end{split}
\end{align}

Tuning the time constants $\tau_{exc}$, and $\tau_{inh}$ and choosing an appropriate $\Delta t$ can often prove to be tedious and challenging. To alleviate this, we introduce a ``learnable" integration step, where $g$ and $h$ are modeled as neural gates. These are computed as specified in Eq.~\ref{si_eq:gates}. $\sigma(.)$ is the sigmoidal function, which squashes activities in the range $[0, 1]$. $\textbf{W}_{g}$, $\textbf{U}_{g}$, $\textbf{W}_{h}$, and $\textbf{U}_{h}$ are convolutional kernels of size $1 \times 1 \times 32 \times 32$.

\begin{align}
\begin{split}
G &= \lb g_{xy}^{(c)} \rb = \sigma (\textbf{W}_{g} * I + \textbf{U}_{g} * Z) \\
H &= \lb h_{xy}^{(c)} \rb = \sigma (\textbf{W}_{h} * E + \textbf{U}_{h} * I) \\
\label{si_eq:gates}
\end{split}
\end{align}

\section{InT \textit{PathTracker}}\label{si_sec:attention}
We visualize InT $A$ attention units on \textit{PathTracker} by simply binarizing the logits, where values greater than $mean(A[t]) + stddev(A[t])$ are set to $1$ and units below that threshold are set to $0$. When applying the same strategy to versions of the InT other than the complete circuit, we found attention that was far more diffuse. For this lesioned InT circuits, adjusting this threshold to be more conservative, choosing two or three or even four standard deviations above the mean, never yielded attention that looked like the complete model. For instance, the closest competitor to the complete InT is one in which its Softplus rectifications are changed to hyperbolic tangents, which remove model constraints for separate and competing forms of Inhibition and Excitation. This model's attention was subsequently diffuse and it also performed worse in generalization than the complete circuit (SI Fig.~\ref{si_fig:attention}).

We also developed a version of the InT with attention that was biased against multi-object tracking. In the normal formulation, InT attention $A$ is transformed with a sigmoid pointwise nonlinearity. This independently transforms every unit in $A$ to be in $[0, 1]$, giving them the capacity to attend to multiple objects at once. In our version biased against multi-object tracking we replaced the sigmoid with a spatial softmax, which normalized the sum of units in each channel of $A$ to 1. This model performed worse than the CNNs or TimeSformer on \textit{Pathtracker} (SI Fig.~\ref{si_fig:extended_benchmark})

\section{InT+TransT}\label{si_sec:tracking}
\begin{figure}[t]
\begin{center}\small
\includegraphics[width=.99\textwidth]{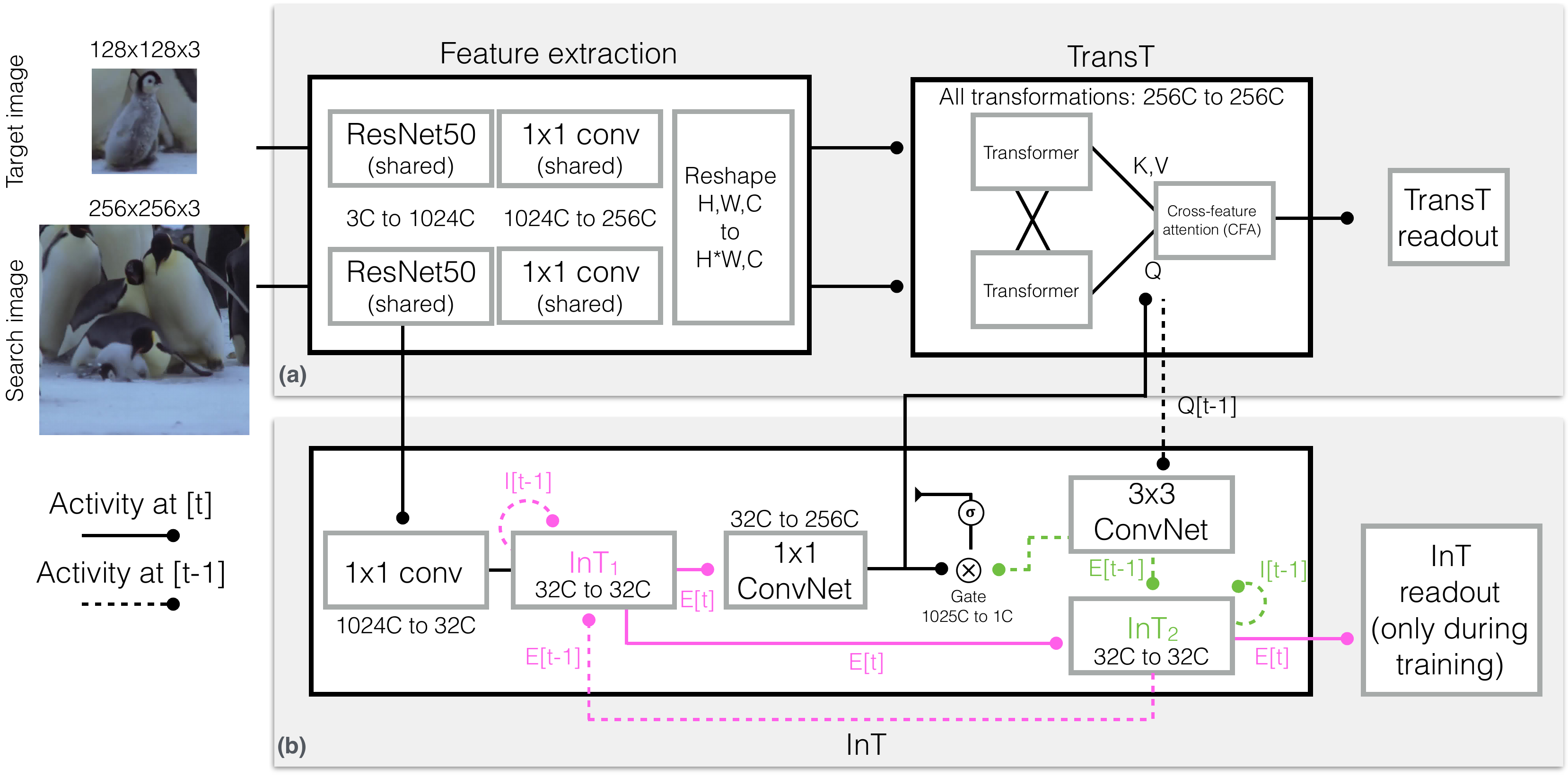}
\end{center}
\vspace{-4mm}
\caption{The (\textit{a}) TransT and (\textit{b}) InT addition to create our the InT+TransT. The InT additively modulate the TransT query (Q) in its CFA, which corresponds to its encoding of the search image which is compared to its encoding of the target. The InT activity is recurrent, and itself modulated by a ``gate'' which captures the similarity of InT activity and the TransT query from the prior step, along with the TransT query entropy. This gate shunts InT activity unless the TransT is low-confidence and the InT and TransT render different predictions, at which point the InT can adjust TransT queries. The InT is further supervised on each step of a video to predict target object bounding boxes.}\vspace{-4mm}
\label{si_fig:inttranst}
\end{figure}

We modify a state-of-the-art tracker, TransT, with our InT circuit, to promote alternative visual strategies for object tracking (Fig.~\ref{si_fig:inttranst}). We note that our InT+TransT model beats almost every benchmark metric on the LaSOT, TrackingNet, and GOT-10K object tracking challenges (SI Table~1).  

\begin{table}[htbp]
\resizebox{\columnwidth}{!}{
\begin{tabular}{|c|c|ccc|ccc|ccc|}
\hline
\label{si_tab:full_tracking}
  Method & Source & \multicolumn{3}{c|}{LaSOT} & \multicolumn{3}{c|}{TrackingNet} & \multicolumn{3}{c|}{GOT-10K} \\
  \cline{3-11} 
   & & AUC & P$_{\textit{Norm}}$ & P & AUC & P$_{\textit{Norm}}$ & P & AO & SR$_{0.5}$ & SR$_{0.75}$ \\
  \hline
  InT+TransT & Ours & \textcolor{red}{\textbf{65.0}} & \textcolor{red}{\textbf{74.0}} & \textcolor{red}{\textbf{69.3}} & \textcolor{red}{\textbf{81.94}} & \textcolor{red}{\textbf{87.48}} & \textcolor{red}{\textbf{80.94}} & \textcolor{blue}{\textbf{72.2}} & \textcolor{blue}{\textbf{82.2}} & \textcolor{red}{\textbf{68.2}} \\
  TransT & CVPR2021 & \textcolor{blue}{\textbf{64.9}} & \textcolor{blue}{\textbf{73.8}} & \textcolor{blue}{\textbf{69.0}} & \textcolor{blue}{\textbf{81.4}} & \textcolor{blue}{\textbf{86.7}} & \textcolor{blue}{\textbf{80.3}} & \textcolor{red}{\textbf{72.3}} & \textcolor{red}{\textbf{82.4}} & \textcolor{red}{\textbf{68.2}} \\
  TransT-GOT & CVPR2021 & - & - & - & - & - & - & {{67.1}} & {{76.8}} & {{60.9}} \\
  SiamR-CNN & CVPR2020 & {{64.8}} & {{72.2}} & - & {{81.2}} & {{85.4}} & {{80.0}} & 64.9 & 72.8 & 59.7 \\
  Ocean & ECCV2020 & 56.0& 65.1& 56.6 & - & - & - & 61.1& 72.1& 47.3 \\
  KYS & ECCV2020 & 55.4 &63.3 &- &74.0 &80.0& 68.8& 63.6& 75.1& 51.5 \\
  DCFST & ECCV2020 & -& - &-& 75.2& 80.9& 70.0& 63.8& 75.3& 49.8 \\
  SiamFC++ & AAAI2020 & 54.4& 62.3& 54.7& 75.4& 80.0& 70.5& 59.5& 69.5& 47.9 \\
  PrDiMP & CVPR2020 & 59.8& 68.8& 60.8& 75.8& 81.6 &70.4& 63.4& 73.8& 54.3 \\
  CGACD & CVPR2020 & 51.8& 62.6& -& 71.1& 80.0& 69.3& -& -& - \\
  SiamAttn & CVPR2020 & 56.0& 64.8& - &75.2& 81.7& -& -& -& - \\
  MAML & CVPR2020 & 52.3& -& -& 75.7& 82.2& 72.5& -& -& - \\
  D3S & CVPR2020 & - &- &-& 72.8& 76.8& 66.4& 59.7& 67.6& 46.2 \\
  SiamCAR & CVPR2020 & 50.7& 60.0& 51.0& -& - &-& 56.9& 67.0& 41.5 \\
  SiamBAN & CVPR2020 & 51.4& 59.8& 52.1& -& -& -& -& -& - \\
  DiMP & ICCV2019 & 56.9 &65.0& 56.7& 74.0& 80.1& 68.7& 61.1& 71.7& 49.2 \\
  SiamPRN++ & CVPR2019 & 49.6& 56.9& 49.1& 73.3& 80.0& 69.4& 51.7& 61.6& 32.5 \\
  ATOM & CVPR2019 & 51.5& 57.6& 50.5& 70.3& 77.1& 64.8& 55.6& 63.4& 40.2 \\
  ECO & ICCV2017 & 32.4& 33.8& 30.1& 55.4& 61.8& 49.2& 31.6& 30.9& 11.1 \\
  MDNet & CVPR2016 & 39.7& 46.0& 37.3& 60.6& 70.5& 56.5& 29.9& 30.3& 9.9 \\
  SiamFC & ECCVW2016 & 33.6& 42.0 &33.9& 57.1& 66.3& 53.3& 34.8& 35.3& 9.8 \\
  \hline
\end{tabular}}
\caption{Object tracking results on the LaSOT~\cite{Fan2018-sb}, TrackingNet~\cite{Muller2018-qn}, and GOT-10K~\cite{Huang2018-yt} benchmarks. First place is in red and second place is in blue. Our InT+TransT model beats all others except for two benchmark GOT-10K scores.}
\end{table}

\paragraph{InT+TransT} We add two InT modules ($InT_1$ and $InT_2$) to the TransT architecture (Fig.~\ref{si_fig:inttranst}). The key difference between these modules and the ones used on \textit{PathTracker} is that they used LayerNorm~\cite{Ba2016-qa} instead of Batch Normalization. This was done because object tracking in natural images is memory intensive and forces smaller batch sizes than what we used for \textit{PathTracker}, which can lead to poor results with Batch Normalization.

$InT_1$ (Fig.~\ref{si_fig:inttranst}b) has the same dimensionality as the one described for \textit{PathTracker} in the main text. ResNet50 features $y \in \mathbb{R}^{1024 \times 32 \times 32}$ are reduced to $z \in \mathbb{R}^{32 \times 32 \times 32}$ by virtue of convolution with a kernel $W_{in} \in \mathbb{R}^{1 \times 1 \times 1024 \times 32}$, i.e., $z = y * W_{in}$. As input, $InT_1$ received $z$. A binary mask $B \in \mathbb{R}^{1 \times 32 \times 32}$ that specified the location of the target object in the very first frame was used to initialize the recurrent excitatory/inhibitory units of $InT_1$. They took values $E_{t=0} = B*W_{E_1}$ and $I_{t=0} = B*W_{I_1}$ respectively, where kernels $W_{E_1},W_{I_1} \in \mathbb{R}^{1 \times 1 \times 1 \times 32}$, and $E_t, I_t \in \mathbb{R}^{32 \times 32 \times 32}$. The subscript $t$ for the recurrent population activities represent an arbitrary time point w.r.t. steps of processing.


To coregister the representations of $InT_1$ and $TransT$, we treat the excitatory units, $E_{t}$, of $InT_1$ by a transformation $f_\phi$ parameterized by three-layer convolutional neural network consisting of $1\times1$ kernels. $f_\phi$ essentially inflates dimensionality, i.e., $f_\phi(E_{t}) \in \mathbb{R}^{256 \times 32 \times 32}$. The network $f_\phi$ had softplus activation functions applied to the output of the first and second layers, and used kernels of dimensions $1\times1 \times32 \times256$, $1\times 1\times 256 \times256$ and $1\times 1\times 256 \times256$ in the three layers respectively. For notational convenience, we refer to $f_\phi(E_{t})$ as $X_t$ in this discussion subsequently.

The ``search frame" query ($Q_t$) for the TransT cross-feature attention (CFA) component was computed as a function of $X_t$ and $Q_{t-1}$ as described here. $Q_{t-1}$ was first subject to a transformation $f_\psi$, parameterized as another convnet, this time to register the query representation to the latent activities of the $InT$ modules. $f_\psi(Q_{t-1}) \in \mathbb{R}^{32 \times 32 \times 32}$ is used to compute two quantities: (a) a measure of spatial certainty in $Q_{t-1}$, and (b) a measure of spatial agreement between $Q_{t-1}$ and $E_t$. For spatial certainty we compute the channel wise $L^2$ norm of $f_\psi(Q_{t-1})$, yielding tensor $H^{(1)} \in \mathbb{R}^{1 \times 32 \times 32}$. For the spatial agreement measure, we compute the feature-wise outer product $H^{(2)} = f_\psi(Q_{t-1}) \otimes E_t \in \mathbb{R}^{1024 \times 32 \times 32}$. The mix-gate $G_{mix}$ was then a convolution on $H = \lb H^{(1)} H^{(2)} \rb$, with a kernel $W_{mix} \in \mathbb{R}^{1 \times 1 \times 1025 \times 1}$, followed by a sigmoidal non-linearity. The final TransT query $Q_t$ was then constructed as the sum of the original $Q_{t}$ and $G_{mix}\odot X_t$. Functionally, this mix-gate helps the InT+TransT compose a hybrid of appearance-free and appearance-based tracker query based on the intrinsic uncertainty of a video frame at a given moment in time. See SI Fig.~\ref{si_fig:inttranst} for a schematic.

The final step in the InT+TransT pipeline is ``top-down'' feedback from the TransT back to $InT_1$. This was done to encourage the two modules to align their object tracks and correct mistakes that emerged in one or the other resource~\cite{Linsley2020-en}. $f_\psi(Q_{t=0})$, computed as described above, was used for initializing the excitatory units of $InT_2$ (InT${_2}$ Fig.~\ref{si_fig:inttranst})b). The inhibitory units of $InT_2$ was initialized with $f_\psi(Q_{t=0}) * W_{I_2}$, where $W_{I_2} \in \mathbb{R}^{1,1,32,32}$. $E_t$ from $InT_1$ served as the input drive to $InT_2$ at every time step $t$. To complete the loop, the recurrent excitatory state of $InT_2$ served as feedback for $InT_1$. We evaluated our InT+TransT on TrackingNet (published under the Apache License 2.0), LaSOT (published under the Apache License 2.0), and GOT-10K (published under CC BY-NC-SA 4.0). See Table~\ref{si_tab:full_tracking} for a full comparison between our InT+TransT and other state-of-the-art models.

\paragraph{Object tracking training and evaluation} The InT+TransT is trained with the same procedure as the original TransT, except that its InTs are given the intervening frames between the target and search images, as described in the main text. Otherwise, we refer the reader to training details in the TransT paper~\cite{Wang2021-no}. Evaluation was identical to the TransT, including the use of temporal smoothing for postprocessing (``Online Tracking''). As was the case for TransT, this involved interpolating the TransT bounding box predictions with a $32\times32$ Hanning window that penalized predictions on the current step $t$ which greatly diverged from previous steps. See~\cite{Wang2021-no} for details.

\end{document}